\pdfoutput=1
\documentclass{article}
\usepackage{float}
\usepackage{amsmath}         %
\usepackage{graphicx}        %
\usepackage{hyperref}        %
\usepackage[utf8]{inputenc}  %
\usepackage{parskip}
\usepackage{makecell}
\usepackage{booktabs}
\usepackage{pifont}
\usepackage{bbm}

\usepackage{multirow}
\usepackage{arydshln}
\usepackage{heatmaps}

\expandafter\def\csname qclabels@csall-plate-nopca-madctrl-plate-nosph@crispr-fs\endcsname{DINOv2 CLS, DINOv2 Patch, SubCell, OpenPhenom, CLOOME, ResNet, {ResNet\\(Untrained)}, CellProfiler}
\expandafter\def\csname qclabels@csall-plate-nopca-madctrl-plate-nosph@crispr-or\endcsname{DINOv2 CLS, DINOv2 Patch, SubCell, OpenPhenom, CLOOME, ResNet, {ResNet\\(Untrained)}, CellProfiler}
\expandafter\def\csname qclabels@csall-plate-nopca-madctrl-plate-nosph@knn-map-compound\endcsname{DINOv2 CLS, CellProfiler, OpenPhenom, DINOv2 Patch, CLOOME, SubCell, ResNet, {ResNet\\(Untrained)}}
\expandafter\def\csname qclabels@csall-plate-nopca-madctrl-plate-nosph@knn-map-tgt2\endcsname{DINOv2 CLS, SubCell, DINOv2 Patch, OpenPhenom, CLOOME, CellProfiler, ResNet, {ResNet\\(Untrained)}}
\expandafter\def\csname qclabels@csall-plate-nopca-madctrl-plate-nosph@knn-recall-compound\endcsname{DINOv2 CLS, DINOv2 Patch, OpenPhenom, CLOOME, SubCell, ResNet, CellProfiler, {ResNet\\(Untrained)}}
\expandafter\def\csname qclabels@csall-plate-nopca-madctrl-plate-nosph@knn-recall-tgt2\endcsname{DINOv2 CLS, DINOv2 Patch, CLOOME, OpenPhenom, SubCell, ResNet, CellProfiler, {ResNet\\(Untrained)}}
\expandafter\def\csname qclabels@csall-plate-nopca-madctrl-plate-nosph@moa-fs-bbbc036-2col\endcsname{DINOv2 CLS, CLOOME, SubCell, DINOv2 Patch, OpenPhenom, CellProfiler, ResNet, {ResNet\\(Untrained)}}
\expandafter\def\csname qclabels@csall-plate-nopca-madctrl-plate-nosph@moa-fs-cpgmoa\endcsname{DINOv2 Patch, DINOv2 CLS, SubCell, CellProfiler, CLOOME, OpenPhenom, {ResNet\\(Untrained)}, ResNet}
\expandafter\def\csname qclabels@csall-plate-nopca-madctrl-plate-nosph@moa-or-bbbc036-2col\endcsname{CLOOME, DINOv2 CLS, SubCell, DINOv2 Patch, OpenPhenom, CellProfiler, ResNet, {ResNet\\(Untrained)}}
\expandafter\def\csname qclabels@csall-plate-nopca-madctrl-plate-nosph@moa-or-cpgmoa\endcsname{DINOv2 CLS, DINOv2 Patch, SubCell, CellProfiler, CLOOME, OpenPhenom, {ResNet\\(Untrained)}, ResNet}
\expandafter\def\csname qclabels@csall-plate-pca64-madctrl-plate-nosph@crispr-fs\endcsname{CLOOME, CellProfiler, DINOv2 Patch, DINOv2 CLS, OpenPhenom, ResNet, SubCell, {ResNet\\(Untrained)}}
\expandafter\def\csname qclabels@csall-plate-pca64-madctrl-plate-nosph@crispr-or\endcsname{CLOOME, CellProfiler, DINOv2 CLS, DINOv2 Patch, OpenPhenom, SubCell, ResNet, {ResNet\\(Untrained)}}
\expandafter\def\csname qclabels@csall-plate-pca64-madctrl-plate-nosph@knn-map-compound\endcsname{CellProfiler, CLOOME, OpenPhenom, DINOv2 CLS, ResNet, DINOv2 Patch, SubCell, {ResNet\\(Untrained)}}
\expandafter\def\csname qclabels@csall-plate-pca64-madctrl-plate-nosph@knn-map-tgt2\endcsname{CellProfiler, DINOv2 CLS, ResNet, DINOv2 Patch, SubCell, OpenPhenom, CLOOME, {ResNet\\(Untrained)}}
\expandafter\def\csname qclabels@csall-plate-pca64-madctrl-plate-nosph@knn-recall-compound\endcsname{CellProfiler, CLOOME, OpenPhenom, DINOv2 CLS, ResNet, SubCell, DINOv2 Patch, {ResNet\\(Untrained)}}
\expandafter\def\csname qclabels@csall-plate-pca64-madctrl-plate-nosph@knn-recall-tgt2\endcsname{CellProfiler, DINOv2 CLS, ResNet, DINOv2 Patch, OpenPhenom, CLOOME, SubCell, {ResNet\\(Untrained)}}
\expandafter\def\csname qclabels@csall-plate-pca64-madctrl-plate-nosph@moa-fs-bbbc036-2col\endcsname{DINOv2 CLS, ResNet, DINOv2 Patch, CellProfiler, OpenPhenom, SubCell, CLOOME, {ResNet\\(Untrained)}}
\expandafter\def\csname qclabels@csall-plate-pca64-madctrl-plate-nosph@moa-fs-cpgmoa\endcsname{ResNet, DINOv2 Patch, DINOv2 CLS, CellProfiler, SubCell, CLOOME, OpenPhenom, {ResNet\\(Untrained)}}
\expandafter\def\csname qclabels@csall-plate-pca64-madctrl-plate-nosph@moa-or-bbbc036-2col\endcsname{DINOv2 CLS, CellProfiler, CLOOME, DINOv2 Patch, ResNet, OpenPhenom, SubCell, {ResNet\\(Untrained)}}
\expandafter\def\csname qclabels@csall-plate-pca64-madctrl-plate-nosph@moa-or-cpgmoa\endcsname{DINOv2 CLS, DINOv2 Patch, ResNet, CellProfiler, SubCell, CLOOME, OpenPhenom, {ResNet\\(Untrained)}}
\expandafter\def\csname qclabels@csall-plate-pca64-madctrl-plate-nosph@negative-control-map-compound\endcsname{CellProfiler, CLOOME, OpenPhenom, ResNet, DINOv2 CLS, DINOv2 Patch, SubCell, {ResNet\\(Untrained)}}
\expandafter\def\csname qclabels@csall-plate-pca64-madctrl-plate-nosph@negative-control-map-tgt2\endcsname{CellProfiler, DINOv2 CLS, DINOv2 Patch, ResNet, OpenPhenom, SubCell, CLOOME, {ResNet\\(Untrained)}}
\expandafter\def\csname qclabels@csall-plate-pca64-madctrl-plate-sphctrl-batch@crispr-fs\endcsname{CLOOME, CellProfiler, DINOv2 CLS, SubCell, ResNet, OpenPhenom, DINOv2 Patch, {ResNet\\(Untrained)}}
\expandafter\def\csname qclabels@csall-plate-pca64-madctrl-plate-sphctrl-batch@crispr-or\endcsname{CLOOME, CellProfiler, SubCell, DINOv2 CLS, OpenPhenom, DINOv2 Patch, ResNet, {ResNet\\(Untrained)}}
\expandafter\def\csname qclabels@csall-plate-pca64-madctrl-plate-sphctrl-batch@knn-map-compound\endcsname{CellProfiler, CLOOME, OpenPhenom, DINOv2 CLS, ResNet, SubCell, DINOv2 Patch, {ResNet\\(Untrained)}}
\expandafter\def\csname qclabels@csall-plate-pca64-madctrl-plate-sphctrl-batch@knn-map-tgt2\endcsname{CellProfiler, ResNet, DINOv2 CLS, DINOv2 Patch, OpenPhenom, SubCell, CLOOME, {ResNet\\(Untrained)}}
\expandafter\def\csname qclabels@csall-plate-pca64-madctrl-plate-sphctrl-batch@knn-recall-compound\endcsname{CellProfiler, OpenPhenom, CLOOME, DINOv2 CLS, ResNet, DINOv2 Patch, SubCell, {ResNet\\(Untrained)}}
\expandafter\def\csname qclabels@csall-plate-pca64-madctrl-plate-sphctrl-batch@knn-recall-tgt2\endcsname{CellProfiler, DINOv2 CLS, ResNet, DINOv2 Patch, OpenPhenom, SubCell, CLOOME, {ResNet\\(Untrained)}}
\expandafter\def\csname qclabels@csall-plate-pca64-madctrl-plate-sphctrl-batch@moa-fs-bbbc036-2col\endcsname{CellProfiler, DINOv2 CLS, ResNet, DINOv2 Patch, SubCell, CLOOME, OpenPhenom, {ResNet\\(Untrained)}}
\expandafter\def\csname qclabels@csall-plate-pca64-madctrl-plate-sphctrl-batch@moa-fs-cpgmoa\endcsname{ResNet, CellProfiler, DINOv2 CLS, DINOv2 Patch, SubCell, CLOOME, OpenPhenom, {ResNet\\(Untrained)}}
\expandafter\def\csname qclabels@csall-plate-pca64-madctrl-plate-sphctrl-batch@moa-or-bbbc036-2col\endcsname{CellProfiler, DINOv2 CLS, CLOOME, ResNet, DINOv2 Patch, OpenPhenom, SubCell, {ResNet\\(Untrained)}}
\expandafter\def\csname qclabels@csall-plate-pca64-madctrl-plate-sphctrl-batch@moa-or-cpgmoa\endcsname{DINOv2 CLS, DINOv2 Patch, CellProfiler, ResNet, SubCell, CLOOME, OpenPhenom, {ResNet\\(Untrained)}}
\expandafter\def\csname qclabels@nocs-pca64-madctrl-plate-nosph@crispr-fs\endcsname{CLOOME, ResNet, DINOv2 CLS, DINOv2 Patch, SubCell, OpenPhenom, {ResNet\\(Untrained)}, CellProfiler}
\expandafter\def\csname qclabels@nocs-pca64-madctrl-plate-nosph@crispr-or\endcsname{ResNet, CLOOME, DINOv2 CLS, DINOv2 Patch, OpenPhenom, SubCell, {ResNet\\(Untrained)}, CellProfiler}
\expandafter\def\csname qclabels@nocs-pca64-madctrl-plate-nosph@knn-map-compound\endcsname{CLOOME, ResNet, CellProfiler, OpenPhenom, DINOv2 CLS, DINOv2 Patch, SubCell, {ResNet\\(Untrained)}}
\expandafter\def\csname qclabels@nocs-pca64-madctrl-plate-nosph@knn-map-tgt2\endcsname{DINOv2 CLS, DINOv2 Patch, ResNet, SubCell, OpenPhenom, CellProfiler, CLOOME, {ResNet\\(Untrained)}}
\expandafter\def\csname qclabels@nocs-pca64-madctrl-plate-nosph@knn-recall-compound\endcsname{CLOOME, CellProfiler, ResNet, OpenPhenom, DINOv2 CLS, DINOv2 Patch, SubCell, {ResNet\\(Untrained)}}
\expandafter\def\csname qclabels@nocs-pca64-madctrl-plate-nosph@knn-recall-tgt2\endcsname{DINOv2 CLS, ResNet, CLOOME, DINOv2 Patch, OpenPhenom, SubCell, CellProfiler, {ResNet\\(Untrained)}}
\expandafter\def\csname qclabels@nocs-pca64-madctrl-plate-nosph@moa-fs-bbbc036-2col\endcsname{DINOv2 CLS, ResNet, SubCell, CLOOME, DINOv2 Patch, OpenPhenom, CellProfiler, {ResNet\\(Untrained)}}
\expandafter\def\csname qclabels@nocs-pca64-madctrl-plate-nosph@moa-fs-cpgmoa\endcsname{DINOv2 Patch, DINOv2 CLS, ResNet, SubCell, CLOOME, OpenPhenom, CellProfiler, {ResNet\\(Untrained)}}
\expandafter\def\csname qclabels@nocs-pca64-madctrl-plate-nosph@moa-or-bbbc036-2col\endcsname{ResNet, DINOv2 CLS, CLOOME, OpenPhenom, DINOv2 Patch, SubCell, CellProfiler, {ResNet\\(Untrained)}}
\expandafter\def\csname qclabels@nocs-pca64-madctrl-plate-nosph@moa-or-cpgmoa\endcsname{DINOv2 CLS, DINOv2 Patch, ResNet, SubCell, CLOOME, OpenPhenom, CellProfiler, {ResNet\\(Untrained)}}

\newcommand{\qcHeatmapLabels}[2]{\csname qclabels@#1@#2\endcsname}

\newcommand{\qcMoaHeatmapBlock}[1]{%
\resizebox{\linewidth}{!}{%
\begin{tikzpicture}
    \begin{scope}
        \heatmapsingle{\textbf{Fraction Significant} \\ BBBC036}
            {NR}{\qcHeatmapLabels{#1}{moa-fs-bbbc036-2col}}{8}{0}{0.150}{3}{data/qc/#1-moa-fs-bbbc036-2col.dat}
    \end{scope}
    \begin{scope}[xshift=1.65cm]
        \heatmapnoy{\textbf{Fraction Significant} \\ cpg-MoA}
            {NR, NSB, NSS}{3}{8}{0}{0.150}{3}{data/qc/#1-moa-fs-cpgmoa.dat}
    \end{scope}
    \begin{scope}[xshift=9.3cm]
        \heatmapsingle{\textbf{Geometric Mean OR} \\ BBBC036}
            {NR}{\qcHeatmapLabels{#1}{moa-or-bbbc036-2col}}{8}{7.7}{12.2}{2}{data/qc/#1-moa-or-bbbc036-2col.dat}
    \end{scope}
    \begin{scope}[xshift=10.95cm]
        \heatmapnoy{\textbf{Geometric Mean OR} \\ cpg-MoA}
            {NR, NSB, NSS}{3}{8}{7.7}{12.2}{2}{data/qc/#1-moa-or-cpgmoa.dat}
    \end{scope}
\end{tikzpicture}%
}%
}

\newcommand{\qcMainMoaHeatmapBlock}[1]{%
\qcMoaHeatmapBlock{#1}%
}

\newcommand{\qcCrisprHeatmapBlock}[1]{%
\resizebox{\linewidth}{!}{%
\heatmappair{%
    \heatmap{\textbf{Fraction Significant} \\ cpg-CRISPR}
        {NR, NSB}{\qcHeatmapLabels{#1}{crispr-fs}}{2}{8}{0}{0.115}{3}{data/qc/#1-crispr-fs.dat}
}{%
    \heatmap{\textbf{Geometric Mean OR} \\ cpg-CRISPR}
        {NR, NSB}{\qcHeatmapLabels{#1}{crispr-or}}{2}{8}{9.8}{15.2}{2}{data/qc/#1-crispr-or.dat}
}%
}%
}

\newcommand{\qcKnnHeatmapBlock}[1]{%
\resizebox{\linewidth}{!}{%
\heatmapgrid{%
    \heatmap{\textbf{kNN Recall@1} --- cpg-tgt2 \\ (301 compounds)}
        {NR, NSB, NSS, NSL}{\qcHeatmapLabels{#1}{knn-recall-tgt2}}{4}{8}{0}{31}{2}{data/qc/#1-knn-recall-tgt2.dat}
}{%
    \heatmap{\textbf{kNN Recall@1} --- cpg-compound \\ (30,138 compounds)}
        {NR, NSB, NSS, NSL}{\qcHeatmapLabels{#1}{knn-recall-compound}}{4}{8}{0}{4.0}{2}{data/qc/#1-knn-recall-compound.dat}
}{%
    \heatmap{\textbf{mAP} --- cpg-tgt2 \\ (301 compounds)}
        {NR, NSB, NSS, NSL}{\qcHeatmapLabels{#1}{knn-map-tgt2}}{4}{8}{0}{5.3}{2}{data/qc/#1-knn-map-tgt2.dat}
}{%
    \heatmap{\textbf{mAP} --- cpg-compound \\ (30,138 compounds)}
        {NR, NSB, NSS, NSL}{\qcHeatmapLabels{#1}{knn-map-compound}}{4}{8}{0}{0.73}{2}{data/qc/#1-knn-map-compound.dat}
}%
}%
}

\newcommand{\qcNegativeControlMapHeatmapBlock}[1]{%
\resizebox{\linewidth}{!}{%
\heatmappair{%
    \heatmap{\textbf{Negative Control mAP} \\ cpg-tgt2}
        {NR, NSB, NSS, NSL}{\qcHeatmapLabels{#1}{negative-control-map-tgt2}}{4}{8}{0}{15}{2}{data/qc/#1-negative-control-map-tgt2.dat}
}{%
    \heatmap{\textbf{Negative Control mAP} \\ cpg-compound}
        {NR, NSB, NSS, NSL}{\qcHeatmapLabels{#1}{negative-control-map-compound}}{4}{8}{0}{2}{2}{data/qc/#1-negative-control-map-compound.dat}
}%
}%
}

\newcommand{\qcPipelineMoaFigure}[3]{%
\begin{figure}[htbp]
\centering
\qcMoaHeatmapBlock{#1}
\caption{MoA enrichment under the #2 normalization variant.}
\label{fig:#3-moa}
\end{figure}
}

\newcommand{\qcPipelineCrisprFigure}[3]{%
\begin{figure}[htbp]
\centering
\qcCrisprHeatmapBlock{#1}
\caption{CRISPR pathway enrichment under the #2 normalization variant.}
\label{fig:#3-crispr}
\end{figure}
}

\newcommand{\qcPipelineKnnFigure}[3]{%
\begin{figure}[htbp]
\centering
\qcKnnHeatmapBlock{#1}
\caption{k-nearest-neighbor replicate retrieval under the #2 normalization variant.}
\label{fig:#3-knn}
\end{figure}
}
\newcommand{\jaccardModelYLabels}{CellProfiler, CLOOME, DINOv2 CLS, DINOv2 Patch, OpenPhenom, ResNet, {ResNet\\(Untrained)}, SubCell}
\newcommand{\jaccardModelXLabels}{CellProfiler, CLOOME, DINOv2 CLS, DINOv2 Patch, OpenPhenom, ResNet, {ResNet (Untrained)}, SubCell}
\newcommand{\jaccardMoaHeatmapBlock}{%
\begingroup
\pgfplotsset{heatmap axis/.append style={x tick label style={font=\scriptsize, rotate=45, anchor=east}}}%
\setlength{\heatmapwidth}{7.4cm}%
\setlength{\heatmapheight}{7.4cm}%
\begin{minipage}[t]{0.52\linewidth}
\centering
\resizebox{\linewidth}{!}{%
    \begin{tikzpicture}
        \heatmap{\textbf{BBBC036 MoA} \\ compounds}
            {\jaccardModelXLabels}{\jaccardModelYLabels}{8}{8}{0}{1}{2}{data/jaccard/bbbc036-moa-jaccard.dat}
    \end{tikzpicture}%
}%
\end{minipage}\hfill%
\begin{minipage}[t]{0.44\linewidth}
\centering
\resizebox{\linewidth}{!}{%
    \begin{tikzpicture}
        \heatmapnoy{\textbf{cpg-MoA} \\ compounds}
            {\jaccardModelXLabels}{8}{8}{0}{1}{2}{data/jaccard/cpg-moa-jaccard.dat}
    \end{tikzpicture}%
}%
\end{minipage}%
\endgroup
}

\PassOptionsToPackage{numbers}{natbib}

\usepackage[eandd, final]{neurips_2026}

\usepackage[utf8]{inputenc} %
\usepackage[T1]{fontenc}    %
\usepackage{hyperref}       %
\usepackage{url}            %
\usepackage{booktabs}       %
\usepackage{amsfonts}       %
\usepackage{nicefrac}       %
\usepackage{microtype}      %
\usepackage{xcolor}         %

\usepackage{pgfplots}
\pgfplotsset{compat=1.18}
\usepgfplotslibrary{colormaps}

\pgfplotsset{
    colormap={ylorrd}{
        rgb255=(255,255,204)
        rgb255=(255,237,160)
        rgb255=(254,217,118)
        rgb255=(254,178,76)
        rgb255=(253,141,60)
        rgb255=(252,78,42)
        rgb255=(227,26,28)
        rgb255=(189,0,38)
        rgb255=(128,0,38)
    },
}

\title{\bmname{}: A Comprehensive Benchmark for Evaluating Representations for Microscopy-Based Morphology Assays}
\author{
  Emre Hayir\\
  Microsoft Research\\
  \texttt{v-emrehayir@microsoft.com} \\
  \And
  Lorin Crawford \\
  Microsoft Research\\
  \texttt{lcrawford@microsoft.com} \\
  \And
  Alex X. Lu \\
  Microsoft Research\\
  \texttt{lualex@microsoft.com} \\
}
\date{\today} %
\usepackage{titlesec}

\usepackage{amssymb}
\titleformat{\paragraph}
{\normalfont\normalsize\bfseries}{\theparagraph}{1em}{}
\titlespacing*{\paragraph}
{0pt}{3.25ex plus 1ex minus .2ex}{1.5ex plus .2ex}
\begin{document}
\newcommand{\bmname}{MorphoHELM}
\newcommand{\todo}[1]{\textcolor{red}{#1}}

\maketitle

\begin{abstract}
Microscopy images contain rich information about how cells respond to perturbations, making them essential to applications like drug screening. To quantify images, researchers often use representation extraction methods, and recent years have seen a proliferation of deep learning methods. While measuring the quality of these representations is essential, evaluation remains fragmented, with each proposed model evaluated on different tasks and datasets, using custom pipelines and metrics, making it difficult to fairly compare models. Here, we introduce \bmname{}, a comprehensive open benchmark for evaluating feature extraction methods for Cell Painting, the most widely-used morphological profiling assay. \bmname{} consolidates evaluation standards in the field, extends and corrects them to be more robust, and evaluates on the widest range of methods to date. A defining feature of the benchmark is that each task is evaluated at different degrees of batch effects (or technical noise), directly quantifying how the ability of methods to detect biological signal degrades as noise increases. Together, these properties enable \bmname{} to detect trade-offs between methods, and we demonstrate that models that excel at certain kinds of biological signal are weaker at others. We show that no existing model outperforms classic computer vision analytic strategies across all settings, which remain the strongest general use-case representations. All datasets, code, and evaluation tools are publicly available at \href{https://github.com/microsoft/MorphoHELM}{https://github.com/microsoft/MorphoHELM}.
\end{abstract}

\section{Introduction}
High-content microscopy has emerged as a key technology for measuring how cells respond to perturbations \citep{cellpainting_intro_Bray2016, a_decade_of_discovery_Seal2024}. Rather than taking a single specialized measurement, microscopy images capture how many different aspects of the morphology of cells may respond to perturbation, from nuclear shape to cytoskeletal organization to organelle distribution \citep{Caicedo2017}. Because cell morphology is the consequence of many biological processes, it serves as a holistic, unbiased readout, making image-based assays critical for drug discovery and functional genomics \citep{Scheeder2018, batch_effects_anne_carpenter}. 

\begin{figure}[t]
\centering
\includegraphics[width=\textwidth]{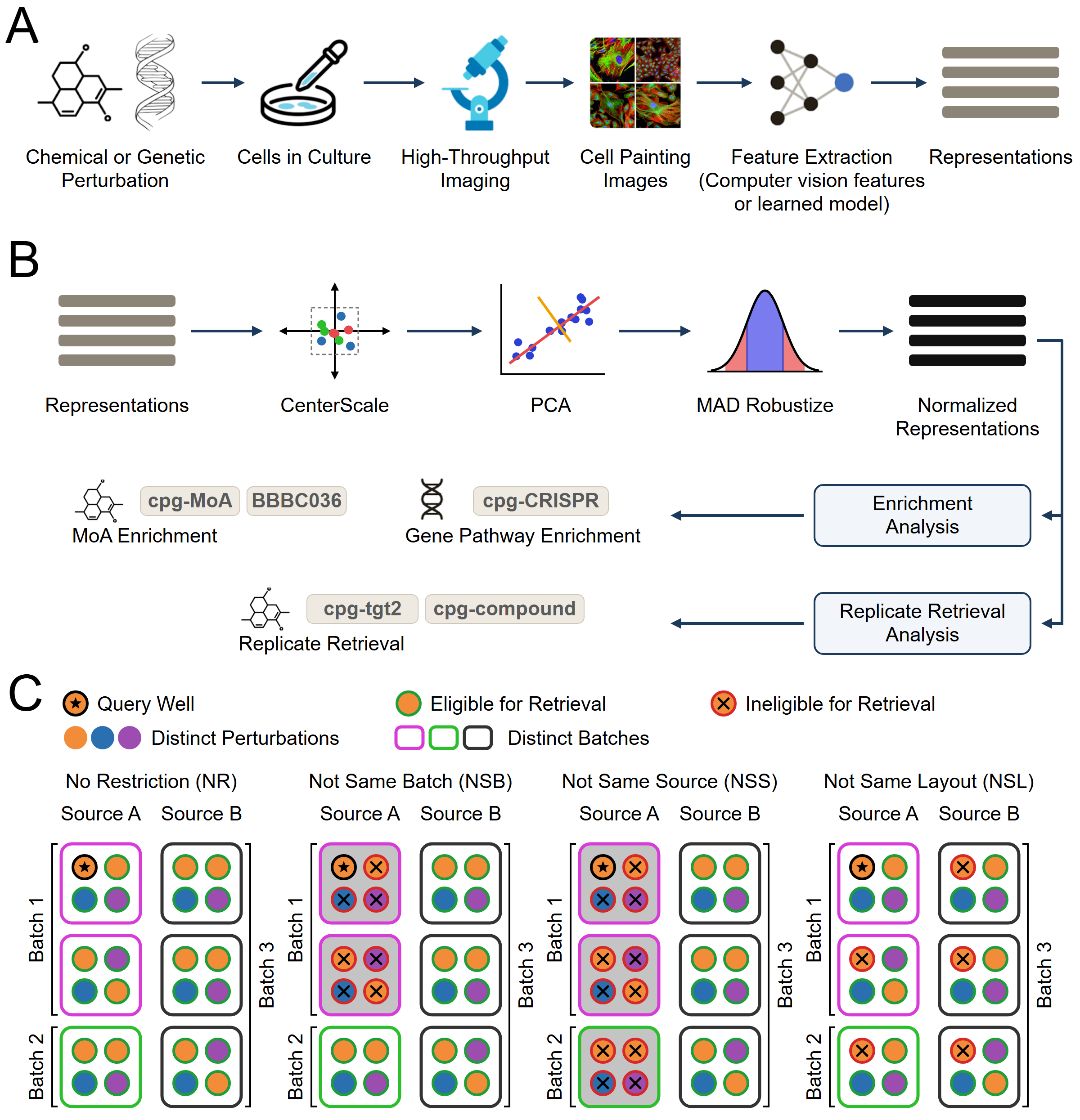}
\caption{Overview figure for MorphoHELM. 
\textbf{(A)} Cell Painting workflow for generating representations. \textbf{(B)} Postprocessing pipeline and benchmark tasks. \textbf{(C)} Batch-effect stringency levels.}
\label{fig:overview}
\end{figure}

Since screens can generate millions of images, their utility depends on the quality of downstream computational  methods. Researchers quantify images using representations, vectors of features that quantify cellular morphology holistically \citep{carpenter_cellprofiler, cellprofiler_v2_Stirling2021}. Because perturbations can alter cell morphology in ways that are statistically reproducible yet visibly imperceptible, analyses demand representations that are not only sensitive to these subtle biological changes but also robust to technical confounders (often referred to as ``batch effects'') such as differences in laboratory-specific imaging hardware \citep{batch_effects_anne_carpenter, arevalo_batch_effects}. This has motivated a growing body of work on representations, spanning classical image analysis pipelines, self-supervised learning, and zero-shot transfer of pretrained vision models \citep{morphological_profiling}.

However, the evaluation of representations remains fragmented. Many works evaluate on their own proposed tasks using different preprocessing pipelines and metrics, thus making direct comparison across studies impossible. While some systematic benchmarks exist \citep{MoA_folds_of_enrichment, recursion_perturbative_maps}, most focus on a narrow range of settings, usually relying on a single definition of biological signal in a single batch effect setting (e.g., requiring methods to compare different experiments reproduced by the same laboratory, but not across laboratories). This means that trade-offs between methods, which can influence adoption for applications, remain unclear. We reason that certain methods can excel for some types of biological signal (e.g., assessing chemical versus genetic perturbations), or identify biological signal in settings where batch effects are less severe.

To address this, we introduce \bmname{} (Holistic Evaluation of Learned Morphological Representations), an open benchmark for evaluating feature extraction methods for the morphological profiling of microscopy images. (Figure~\ref{fig:overview}). Here, we focus on Cell Painting data, the most widely adopted assay in the field, which stains cells with six fluorescent dyes imaged over five channels \citep{cellpainting_intro_Bray2016} to visualize eight cellular compartments. The scalability of the Cell Painting protocol has made it an important tool in drug discovery, supporting mechanism-of-action identification, toxicity prediction, and hit prioritization \citep{a_decade_of_discovery_Seal2024}.\bmname{} provides several key contributions:
\begin{itemize}
    \item We aim to be more comprehensive of biological signal. We aggregate existing proposals for evaluating Cell Painting representations, including enriching for chemical compounds with similar mechanisms of action (MoA) during retrieval, enriching for genes of similar function during retrieval, and clustering replicates of compound perturbations.
    \item Although our tasks are inspired by previous benchmarks, we evaluate a broader range of representation methods than prior studies. We also identify and address implementation choices that can distort benchmark interpretation, including clustering metrics that are sensitive to latent dimensionality and correlation structure, and odds-ratio summaries that can be skewed by biased imputation of infinite values. We show that these issues can substantially affect conclusions about representation quality.
    \item Finally, we extend tasks so that they are assessed with varying degrees of batch effects, quantifying how biological signal degrades as confounding becomes more severe. This enables the articulation of trade-offs between methods as a function of batch effect severity.
\end{itemize}

We evaluate a diverse set of feature extraction methods adopted in the field: classical hand-crafted features, transfer learning from vision foundation models, and domain-specific models pre-trained on large-scale microscopy data. All datasets, code, and evaluation tools are publicly available at \href{https://github.com/microsoft/MorphoHELM}{https://github.com/microsoft/MorphoHELM}.

\section{Related Works}
\textbf{Evaluating sensitivity to biological signal:} A variety of metrics have been proposed to assess if representation methods are sensitive to morphological changes induced by a perturbation. Some works rely upon replicate retrieval or clustering \citep{compound_replicate_chammi, rxrx1-dataset}, which tests if replicates of the same perturbation are grouped in representation space. While this strategy rewards methods that identify reproducible biological changes across independent experiments, it can be noisy as some perturbations may not induce morphological responses from cells. Hence, other works have proposed assessing if biologically related perturbations\textemdash MoA for compound perturbations \citep{MoA_folds_of_enrichment}, and gene pathway or interaction for genetic perturbations \citep{recursion_perturbative_maps}\textemdash are preferentially retrieved given a sample. However, these strategies are limited by the availability and bias of labels. Hence, existing evaluations of biological signal have trade-offs. For this reason, we aggregate and standardize all of these strategies in \bmname{}, while prior benchmarks have typically focused on one strategy.

\textbf{Testing robustness to batch effects:} Prior replicate retrieval and clustering benchmarks test at different levels of batch effect severity. The earliest benchmarks using the BBBC021 dataset originally proposed reporting Not Same Batch (NSB) matching, which restricts nearest neighbors to those not in the same experimental batch \citep{bbbc021_dataset, Ando_bbbc021}. This idea has been adopted in later works, most ambitiously by the benchmarks of \citet{arevalo_batch_effects}, which evaluate five scenarios of increasing complexity. The importance of this procedure was demonstrated with the RxRx1 \citep{rxrx1-dataset} benchmark, which originally did not implement this and reported accuracies of over 80\% on gene perturbation replicate retrieval. Updated work in the RxRx1-WILDS \citep{rxrx1_wilds} version of the benchmark with held out batches resulted in accuracies around 40\%, suggesting performance was overestimated due to overfitting on batch effects. Despite their importance, these ideas have not yet been implemented for MoA or gene pathway enrichment, so \bmname{} extends batch effect robustness testing to these strategies.

\section{Methods}

\subsection{Overview of Benchmark Design}
\bmname{} includes three categories of biological signal retrieval tasks: 
\begin{itemize}
    \item \textbf{Mechanism of action (MoA) enrichment} measures if, given images of cells treated with a chemical compound, a method retrieves images of cells treated with other compounds with the same MoA.
    \item \textbf{Gene pathway enrichment} measures if, given images of cells with a specific gene knocked out, a method retrieves images of cells with other gene knock-outs where the genes have similar functions (e.g., are in the same pathway or interact).
    \item \textbf{Replicate retrieval consistency} measures if, given an image of a cell treated with a perturbation, a method preferentially retrieves replicates treated with the same perturbation (as opposed to other, different perturbations). 
\end{itemize}

A central design principle of \bmname{} is that biological signal retrieval and robustness to batch effects cannot be meaningfully evaluated in isolation. A model that is effective at comparing images generated by the same laboratory may fail entirely comparing images across laboratories. \bmname{} evaluates each biological retrieval task across different levels of batch effect stringency, directly quantifying how signal degrades as technical variation increases.
We define four levels of stringency that apply across all tasks (although note some tasks may be missing some levels due to data availability):
\begin{enumerate}
\item \textbf{No Restriction (Baseline/NR)}: Retrieved samples can be any sample.
\item \textbf{Not Same Batch (NSB)}: Candidates from the same experimental batch are excluded, testing robustness to day-to-day variation between batches.
\item \textbf{Not Same Source (NSS)}: Candidates from the same institution are excluded, testing robustness to domain shifts like microscopy hardware differences.
\item \textbf{Not Same Layout (NSL)}: High-throughput microscopy experiments are conducted on plates, which contain hundreds of individual wells, each containing an independent perturbation. In many experiments, the position of perturbations on plates is fixed across replicates, but previous work has demonstrated that position on a plate induces significant batch effects \citep{batch_effects_anne_carpenter}. Hence, in this setting, we exclude candidates sharing the same position.
\end{enumerate}

A robust model minimizes the performance drop across levels, while a biologically sensitive model may have the highest performance across all or some of these settings. More details on the implementation of these levels of batch effect stringency are provided in \hyperref[app:stringency]{Appendix A.1}.

\subsection{Metrics}

\subsubsection{Enrichment task metrics}
Our MoA and gene pathway enrichment tasks rank image representations based upon their similarity to a query sample (by cosine similarity) and measure if samples sharing the same biological label are enriched at the top of the ranking. Previous works have proposed different metrics across tasks. For MoA enrichment, \citet{MoA_folds_of_enrichment} proposed reporting the average odds ratio of enrichment across samples. For gene pathway enrichment, \citet{recursion_perturbative_maps} proposed the percentage of samples with a significant number of retrievals sharing the same gene pathway at a predefined cut-off. We reasoned these metrics complement each other: the average odds ratio can be skewed by a small number of highly enriched samples, while the percent significant may overly reward methods that have low recall but significant enrichment across many samples. Hence, we report both metrics for both tasks, harmonizing the evaluations. A more detailed description of both metrics is available in \hyperref[app:moa-metric]{Appendix A.2} and \hyperref[app:pathway]{Appendix A.3}.

Additionally, we update the average odds ratio metric to be more robust to biological labels with few members. In these cases, there is a higher chance of all members being retrieved within the top of the ranking, yielding an infinite odds ratio. \citet{MoA_folds_of_enrichment} arbitrarily impute with the total sample size when this occurs which results in a tail of extremely high odds ratios that skews the average (\hyperref[app:haldane]{Appendix D.1}). Instead, we apply modified Haldane-Anscombe correction \citep{modified_haldane_anscombe}, which is less arbitrary and quenches extreme values (\hyperref[app:haldane]{Appendix D.1}). Finally, as some outliers still persist, we report the geometric mean as our final metric, which further reduces the influence of extreme outliers as shown in statistical practice \citep{log_odds_ratio}. 

\subsubsection{Replicate retrieval consistency metrics}
We adopt K-nearest-neighbor (kNN) accuracy metrics introduced in a variety of previous works \citep{Ando_bbbc021, micon}. Given query samples treated with specific perturbations, this metric measures the percent of times the nearest neighbor (based upon cosine similarity in representation space) is treated with the same perturbation when retrieving from a large database of samples treated with different perturbations. 

However, kNN is a highly local metric that may not fully identify methods that globally cluster samples with the same perturbation. To address this, \citet{arevalo_batch_effects} proposed applying clustering metrics originally adopted by single-cell transcriptomic benchmarks, including Leiden ARI, Leiden NMI, and Silhouette label scores. We found that these metrics are brittle when the number of unique perturbations is large relative to the number of replicates per perturbation, as is the case in large-scale compound screens. Specifically, in the scenario of \citet{arevalo_batch_effects} that evaluates across multiple microscopes, laboratories, and tens of thousands of compounds with few replicates, the resulting clustering problems can have over 80,000 clusters with as few as 2 samples per cluster. Under these conditions, clustering metric differences between methods are not driven by biological signal: in \hyperref[app:clustering]{Appendix D.3}, we show that randomly permuting treatment labels while preserving batch structure yields the same performance gaps between methods, indicating the gaps are attributable to differences in latent dimensionality and correlation structure rather than biological content. \citet{arevalo_batch_effects} propose contrasting these clustering metrics for the same representations on biological labels versus batch labels, to assess trade-offs in biological sensitivity versus batch effects. \bmname{} instead evaluates replicate retrieval at increasing degrees of batch effects.

We also instead adopt as our global metric, the mean average precision (mAP) across samples. We rank other samples based on their distance from a query sample. The mAP then measures if samples with the same compound treatment are preferentially retrieved at the top of the ranking. However, including all samples means the mAP focuses on measuring if methods identify morphological impacts of compounds that distinguish them from other compounds. Hence, we also calculate the mAP using only negative controls as confounders, which less stringently measures if methods detect whether cells exhibit some phenotype distinct from negative controls. For completeness, we also report the mean F1 score among the $k$ nearest neighbors (F1@$k$), which balances precision and recall of replicate retrieval; model rankings under F1@$k$ closely agree with those under mAP (\hyperref[app:f1]{Appendix D.6}). A full description of how metrics are implemented across levels of batch effect severity is in \hyperref[app:stringency]{Appendix A.1}.

\subsection{Dataset Curation}

As our benchmarks rely on identifying datasets with chemical and genetic perturbations, and with replicates reproduced across batches, laboratories, and plate layouts, we curated public datasets derived from two primary sources: the Broad Bioimage Benchmark Collection (BBBC) and the JUMP-Cell Painting (JUMP-CPG) consortium. A summary of these datasets is provided in Table \ref{tab:datasets}, with detailed statistics in \hyperref[tab:dataset_details]{Appendix B.1}.

\begin{table}[h] 
    \centering
    \caption{Summary of the included datasets used for evaluation.}
    \label{tab:datasets}
    \renewcommand{\arraystretch}{1.2} 
    \begin{tabular}{lccc}
        \toprule
        \textbf{Dataset Name} & \textbf{\makecell{Replicated Across\\Multiple Institutions}} & \textbf{\makecell{Genetic\\Perturbation}} & \textbf{\makecell{Chemical\\Perturbation}} \\
        \midrule
        BBBC036 & 
        \ding{55} & \ding{55} & \ding{51} \\
        cpg-MoA & 
        \ding{51} & \ding{55} & \ding{51} \\
        cpg-CRISPR & 
        \ding{55} & \ding{51} & \ding{55} \\
        cpg-target2 & 
        \ding{51} & \ding{55} & \ding{51} \\
        cpg-compound & 
        \ding{51} & \ding{55} & \ding{51} \\

        \bottomrule
    \end{tabular}
\end{table}

For gene set enrichment tasks, we follow \citet{recursion_perturbative_maps} and use labels derived from five databases of cellular signaling pathways and protein complexes: CORUM \citep{CORUM}, HuMAP \citep{humap}, StringDB \citep{StringDB}, SIGNOR \citep{SIGNOR}, and Reactome \citep{Reactome}. These databases are described further in Appendix~\ref{app:databases}. For MoA enrichment tasks, we use MoA labels for BBBC036 and cpg-MoA compounds, originally transferred from the Drug Repurposing Hub \citep{drug_repurposing_hub}.

\subsection{Model Selection}
We curate a cross-section of representation learning and extraction strategies currently used for morphological profiling including classical feature engineering methods (CellProfiler \citep{carpenter_cellprofiler}), transfer learning from natural vision foundation models (ImageNet-Trained ResNet101 \citep{resnet}, DINOv2 (ViT-B/14) \citep{dinov2}), and microscopy foundation models trained on large-scale Cell Painting datasets (OpenPhenom \citep{OpenPhenom}), on distinct fluorescent microscopy datasets that have been demonstrated to transfer well to Cell Painting datasets (SubCell \citep{subcell}), and multi-modally on Cell Painting and chemical compound structures (CLOOME \citep{cloome}). Since it remains undetermined if using a CLS token embedding or averaging patch embeddings is best practice, we experiment with both when performing transfer learning from DINOv2,. In addition, we include an untrained, randomly initialized ResNet101 as a lower bound baseline. A detailed description of models is available in \hyperref[app:models]{Appendix C.1}. All models are used frozen, without fine-tuning, consistent with how they are deployed in practice \citep{OpenPhenom, subcell}. 

\subsection{Preprocessing, Postprocessing and Quality Control Pipelines}

To ensure reliability, we implemented a rigorous quality-control pipeline. This pipeline addresses technical artifacts at the image level and filters low-quality samples to maximize the chance that samples reflect their label, as opposed to having no signal identifiable by any method (e.g., because they are empty images). In addition, we preprocess images and postprocess representations consistently to ensure fairness across all methods.

\subsubsection{Image Preprocessing}
Prior to feature extraction, raw microscopy images were converted to 8-bit PNG format to standardize dynamic range and reduce storage requirements. We then applied illumination correction to all fields of view to rectify uneven lighting and vignetting artifacts introduced by the optical system.

\subsubsection{Sample Quality Control}
We applied a series of filtering steps to remove noisy data at the batch, well, and channel levels:

\begin{itemize}
    \item \textbf{Channel De-duplication:} Cell Painting collects five fluorescent channels for each microscopy image. As each channel should show a distinct set of cellular compartments, images with duplicated imaging channels are removed as artifacts.
    \item \textbf{Cell Count Filtering:} We removed wells with insufficient cell counts. We estimated the global cell count distribution by sampling 5,000 images. Based on this distribution, we removed wells falling in the bottom 5th percentile.
\end{itemize}

\subsubsection{Feature Normalization}
While some previous benchmarks have used representations as-is (e.g. \citep{MoA_folds_of_enrichment}), we observed that this does not control for varying dimensionality or redundant features across methods. In particular, CellProfiler features depend upon scaling and PCA for performance {\hyperref[app:cellprofiler-dim]{Appendix D.4}}, likely due to its much higher dimensionality and inclusion of many redundant features. Hence, we decided to standardize all features to the same dimensionality and postprocess with basic batch effect correction methods in a fixed feature normalization pipeline applied to each method:

\begin{enumerate}
    \item \textbf{Platewise Scaling and PCA:} First, features were scaled to a standard range on a per-plate basis. We then applied Principal Component Analysis (PCA) across the entire dataset to reduce dimensionality and denoise the feature space. This global fit matches our aim of integrating an existing multi-institution dataset in which all data are available; when new data must be analyzed with previously fitted components, PCA should instead be fit only on seen data. For our benchmark, we observe that fitting PCA on only the query source yields nearly identical results and unchanged model rankings in the settings we evaluated (\hyperref[app:unleaked-pca]{Appendix D.5})
    \item \textbf{Platewise Robust Standardization:} To align distributions across different experimental runs, we applied a Median Absolute Deviation (MAD) "Robustize" transformation. This was performed on a per-plate basis.
\end{enumerate}

Previous works have proposed applying whitening transformations to reduce batch-specific signatures in representations. However, these works generally assess whitening as part of a pipeline with scaling and PCA, and do not isolate the impact of steps. We systematically ablate parts of this pipeline, and find that scaling and PCA improve representation performance across tasks and settings, but whitening harms performance in most cases \hyperref[app:cellprofiler-dim]{Appendix D.4}. This is consistent with findings of \citet{arevalo_batch_effects} where whitening can reduce performance under high noise scenarios. We also show the above proposed pipeline yields performance increases for almost all representations and settings. 

\section{Results}
\subsection{MoA Enrichment}

\begin{figure}[htbp]
\centering
\qcMainMoaHeatmapBlock{csall-plate-pca64-madctrl-plate-nosph}
\caption{MoA enrichment. Restrictions: NR = No Restriction, NSB = Not Same Batch, NSS = Not Same Source.}
\label{fig:moa_bbbc036}
\end{figure}

We report MoA enrichment results in Figure \ref{fig:moa_bbbc036} where we observe that both metrics degrade in performance as batch effects become more severe. All models perform best on settings with no matching restriction (we note that the BBBC036 dataset only has this setting due to the nature of its data collection). However, all models also exhibit the worst performance on the ``Not Same Source'' setting. This performance drop is particularly stark for the Fraction Significant metric. Statistically, this metric has the same interpretation as a false discovery rate equal to 0.05. All models perform below this level, suggesting that retrieving biologically similar compounds when comparing with data generated from a different distribution remains an unsolved problem, even when institutions use similar collection procedures and when data is postprocessed to mitigate batch effects. We assess the robustness of model rankings using confidence intervals from resampling sources of technical variation and an increased number of permutations in \hyperref[app:stats]{Appendix D.7}.

Across all settings, natural image foundation models are the best performers, with DINOv2 CLS token outperforming ResNet, possibly due to its self-supervised training strategy enabling a more expansive training dataset. Foundation models trained on microscopy images perform worse than DINOv2. This is most pronounced on the Fraction Significant metric, and less so on the Geometric Mean OR.

We observed some disagreement between our Fraction Significant and Geometric Mean OR metrics: for the ``Not Same Batch'' setting, while DINOv2 and ResNet outperform CellProfiler and CLOOME for Fraction Significant, with the exception of DINOv2 using the CLS token, these models perform worse for Geometric Mean OR. This suggests that, while natural image foundation models may capture a wider breadth of MoAs, other models may more deeply enrich for particular MoAs. To investigate this behavior further, we calculated the Jaccard similarity between lists of compounds that retrieve significantly enriched MoAs as shown in Figure \ref{fig:jaccard-overlap-moa}. We find methods exhibit trade-offs in which compounds they discover enriched signal. Crucially, while natural image foundation models enrich for similar compounds across models, other strategies find associations for distinct compounds not uncovered by natural image foundation models. This result cautions against drawing the conclusion that because they achieve the highest metrics on our benchmark, natural image foundation models are superior for all applications, as they appear to be relatively insensitive to certain MoAs captured by other models.

\begin{figure}[h]
\centering
\jaccardMoaHeatmapBlock
\caption{Pairwise Jaccard similarity between least-restrictive MoA significant-compound discovery sets under the main postprocessing setting.}
\label{fig:jaccard-overlap-moa}
\end{figure}
\subsection{Gene Pathway Enrichment}
\begin{figure}[htbp]
\centering
\qcCrisprHeatmapBlock{csall-plate-pca64-madctrl-plate-nosph}
\caption{Gene Pathway Enrichment.}
\label{fig:crispr}
\end{figure}

We report average cpg-CRISPR metrics over all gene annotation databases in Figure \ref{fig:crispr} (results for individual databases are available in {\hyperref[app:gene_path]{Appendix D.2}}). Similar to MoA enrichment, performance drops as the severity of batch effects increases. Note that due to the nature of the cpg-CRISPR dataset, only the Not Same Batch setting is possible.

Compared to the MoA enrichment benchmarks, the performance of models changes. Although the natural image foundation models (DINOv2 and ResNet) remain strong performers, CellProfiler and CLOOME emerge as the top models in the No Restriction setting (although CLOOME loses its performance advantage in the Not Same Batch setting). Conversely, ResNet is the worst performing model on our Geometric Mean OR metric (except for our untrained ResNet baseline), despite being relatively strong for MoA enrichment. This indicates that different models may have trade-offs across biological tasks. Intriguingly, models may exhibit generalization among tasks that doesn't match its training distribution: CLOOME is trained multi-modally on chemical compounds, which are more relevant for MoA enrichment than genetic function enrichment.

\subsection{Replicate Retrieval Consistency}
\begin{figure}[htbp]
\centering
\qcKnnHeatmapBlock{csall-plate-pca64-madctrl-plate-nosph}
\caption{kNN replicate retrieval.}
\label{fig:knn}
\end{figure}

We show metrics for replicate retrieval consistency in Figure~\ref{fig:knn}. As with prior tasks, performance degrades across models with stronger batch effects. For these datasets, we have some replicates employing different plate layouts on which all models perform substantially worse relative to even the Not Same Source evaluations for the cpg-target2 dataset, suggesting that position on a plate is a major contributor to batch effects (consistent with observations previously documented by \citet{cpg0016_dataset}). This observation is not made for the cpg-compound dataset. We speculate that this is because this dataset reflects a more challenging replicate retrieval problem due to the much larger number of compounds, so even at the Not Same Source level, all methods already have very low performance. However, for the kNN Recall@1 metric, models show some improvements over random matching. For example, the expected random Recall@1 for cpg-target2 is 0.33, and the weakest models show about 10$\times$ better performance even in the most challenging Not in Same Layout setting, contrary to MoA enrichment, suggesting for replicate retrieval consistency, all methods detect some reproducible biological signature.

CellProfiler features are superior to any representation learning strategy on the kNN Recall@1 and mAP metrics across most settings. This contradicts some previous works that show CellProfiler features being outperformed by representation learning procedures \citep{MoA_folds_of_enrichment}. One explanation is in this work, we apply standardized postprocessing operations that control for dimensionality, de-correlate redundant features, and reduce batch effects to all representations. While this procedure benefits all methods, it most disproportionately benefits CellProfiler features ({\hyperref[app:cellprofiler-dim]{Appendix D.4}}), which are higher dimensionality and contain many more redundant features than representations learned by deep learning. Additionally, we note, unlike the MoA and gene pathway enrichment tasks which query using averaged representations across multiple samples and can reduce noise, replicate retrieval consistency operates on individual samples, so it is possible that CellProfiler features are more robust in this setting.

\section{Conclusion}

We introduce \bmname{}, a comprehensive benchmark for evaluating representations for microscopy-based morphology assays. While \bmname{} draws from existing proposals for evaluation, it aggregates strategies from previously proposed benchmarks in a single standardized benchmark suite, corrects issues in the implementation of previous proposals, and evaluates over a much wider range of methods than previous works. Each of these advances yields insights for the field.

First, by contrasting evaluation strategies, \bmname{} exposes trade-offs between methods. We show trade-offs with microscopy foundation models. For example, CLOOME leads unrestricted gene pathway enrichment but drops sharply across sources; SubCell is weak there but more robust across sources. CellProfiler features perform consistently well across all tasks, which may be consistent with continued adoption of these features despite representation learning methods frequently claiming better performance. Although a ranking of methods by performance emerges for each task, methods still exhibit trade-offs in what kinds of biological signal they detect within tasks: for our MoA enrichment tasks, we show that distinct methods enrich MoAs for distinct compounds.

Second, correcting implementation issues and standardizing processing alters interpretation. Most notably, we show that CellProfiler performance may have been underestimated in previous works, because lightweight postprocessing corrections like PCA disproportionately affect these features which can be seen in \hyperref[app:cellprofiler-dim]{Appendix D.4}. 

Third, by benchmarking a wider range of models, we show that microscopy foundation model efforts fall short of using frozen natural image foundation models out-of-the-box, with DINOv2 outperforming all microscopy foundation models we evaluated in most settings. This suggests a trade-off between natural vision and microscopy models. Natural vision models may provide a greater diversity of generic features \citep{dinov2}, whereas microscopy foundation models seem to compromise this breadth despite being trained on domain specific images. This is consistent with the previous works that show that natural image models are competitive with domain specific ones \citep{OpenPhenom}. These results also inform how to choose representations in practice. No single model performs best across tasks, and the choice of model should be adjusted to the task and the data source. Microscopy foundation models perform strongly on benchmarks whose data closely resemble their training distribution \citep{OpenPhenom}, but fail to improve over natural image models across sources.

Together, these insights mean that \bmname{} highlights opportunities for model improvement. When comparing data generated by different laboratories, we showed that models have negligible performance over random guessing. This suggests that detecting biological signal in a high noise environment remains an area for significant improvement in the future.

We note some limitations and future directions. We apply a simple, standardized normalization pipeline that ensures that no method receives bespoke treatment. However, some prior methods implement extensive postprocessing of their features \citep{Caicedo2017, MoA_folds_of_enrichment, arevalo_batch_effects}, and we show that these choices can change model rankings, although our main conclusions hold when each model is evaluated under its best pipeline (\hyperref[app:cellprofiler-dim]{Appendix D.4}). Benchmarks should therefore report postprocessing as part of their specification, and future work should assess whether some postprocessing strategies are more critical to some methods than others. 

Critically, we showed that none of the representation learning methods outperform CellProfiler in the replicate retrieval consistency task, and unlike the other two tasks, this task operates on individual sample-level representations (rather than averaging multiple samples together). Averaging over samples reduces noise because variability in representations across replicates averages out, so it is possible this task requires more extensive post-processing than others. Furthermore, CellProfiler features were extracted from 16-bit images, whereas learned models received 8-bit images. We expect the impact of this to be negligible relative to other factors we investigate, as most computational analyses of Cell Painting use 8-bit images \citep{cpg0016_dataset, MoA_folds_of_enrichment}, which has been argued to improve robustness by filtering out imaging artifacts \citep{odje2024}, and the pretrained models we benchmark are themselves trained on 8-bit images, so they are unlikely to exploit the additional bit depth.

Additionally, we only focus on Cell Painting, but morphological profiling extends to other staining panels, label-free imaging, and live-cell assays. The transferability of our findings to these other modalities is unverified. Extending the benchmark to additional assays would test whether the conclusions reflect more general properties of the methods.
\section{Broader Impacts}
Cell Painting benchmarks can accelerate biological discovery by making representation-learning methods easier to compare across datasets, perturbation types, and batch-effect settings. By emphasizing robustness under increasingly stringent evaluation conditions, this work may help reduce overoptimistic claims and encourage more reliable deployment of image-based profiling methods in drug discovery and functional genomics. At the same time, benchmarks can shape research incentives: if used uncritically, they may overemphasize leaderboard performance over biological validity, reproducibility, or downstream utility. We encourage users to use \bmname{} results in accordance with dataset provenance, perturbation coverage, and task-specific biological assumptions.
\newpage

\begin{ack}
This work was funded by Microsoft Research. Lorin Crawford and Alex X. Lu are employees of Microsoft and hold Microsoft stock. Emre Hayir was employed by Microsoft during this work. The authors declare no other competing interests.
\end{ack}

\bibliographystyle{unsrtnat}       %
\bibliography{references}

\newpage
\section*{Appendix}
\setcounter{figure}{0}
\setcounter{table}{0}

\renewcommand{\thefigure}{\arabic{figure}}
\renewcommand{\thetable}{\arabic{table}}

\renewcommand{\figurename}{Appendix Figure}
\renewcommand{\tablename}{Appendix Table}
\subsection*{Appendix A: Tasks and Metrics}

\subsubsection*{A.1 Batch Effect Stringency Levels}
\label{app:stringency}

\bmname{} evaluates each biological retrieval task across four levels of batch effect stringency. The levels are defined consistently across tasks, but their implementation differs depending on whether the task operates on consensus profiles (enrichment tasks) or individual well-level profiles (replicate retrieval). The four levels are:
\begin{enumerate}
    \item \textbf{No Restriction (Baseline/NR)}: Retrieved samples can be any sample.
    \item \textbf{Not Same Batch (NSB)}: Candidates from the same experimental batch are excluded, testing robustness to day-to-day variation between batches.
    \item \textbf{Not Same Source (NSS)}: Candidates from the same institution are excluded, testing robustness to domain shifts such as microscopy hardware differences.
    \item \textbf{Not Same Layout (NSL)}: Candidates sharing the same plate position as the query are excluded, controlling for position-induced batch effects \citep{batch_effects_anne_carpenter}.
\end{enumerate}

For the enrichment tasks (MoA and gene pathway enrichment), stringency is implemented by constructing separate consensus profiles from non-overlapping subsets of replicates. At the No Restriction level, all replicates of a perturbation are aggregated into a single consensus profile. At stricter levels, replicates are split by the relevant experimental condition: by experimental batch (NSB) or by institution (NSS). The query consensus is built from each unique batch. For NSB, query consensus is compared against any other batchwise representations of compounds and for NSS query consensus is compared against batchwise representations that are not in the same source as the query batchwise representation. In cases where a compound is replicated in more than one batch we use majority voting and if more than half of the batchwise representations of the compound are significantly enriched we consider the compound enriched. For each batchwise representation top 1\% of the unique compound representation are considered. 

It is also important to note that the No Restriction setting for enrichment tasks have more replicates available for profile aggregation compared to NSB and NSS stringencies. This might create more robust biological profiles in the No Restriction setting in addition to the more lenient batch effect environment.

For replicate retrieval, stringency is implemented differently because the task operates on individual well-level profiles rather than consensus representations. Rather than constructing separate consensus profiles, we apply exclusion constraints to the neighbor search: at each stringency level, candidates sharing the relevant experimental condition with the query are removed from consideration prior to ranking.

Not all stringency levels apply to all datasets. NSS cannot be evaluated on single-institution datasets (BBBC036, cpg-CRISPR) because no cross-institutional candidates exist. The available levels per dataset are noted in the task descriptions below.

\subsubsection*{A.2 MoA Enrichment}
\label{app:moa-metric}

This task quantifies a model's ability to cluster compounds sharing the same Mechanism of Action. We average well-level profiles into a consensus compound representation to mitigate outlier noise, then rank candidate compounds by cosine similarity to the query consensus.

We evaluate this task on two datasets:
\begin{itemize}
    \item \textbf{BBBC036}: A single-institution dataset from the Broad Institute. MoA retrieval within this dataset serves as an upper-bound estimate of performance in the absence of cross-site batch effects. Only the No Restriction level is evaluated.
    \item \textbf{cpg-MoA}: A subset of the JUMP-Cell Painting dataset containing chemical perturbations with mechanism-of-action annotations. Compounds are profiled across multiple batches and institutions, enabling MoA retrieval to be evaluated under No Restriction, Not Same Batch, and Not Same Source stringency levels.
\end{itemize}

\paragraph{Metric: Permutation-Based Enrichment Analysis.}
We use a cutoff $k$ to convert the ranked list of candidate compounds into a retrieval set. Let $a$ be the number of samples matching the same MoA in the retrieval set, $b$ the number of non-matching samples in the retrieval set, $c$ the number of matching samples not retrieved, and $d$ the number of non-matching samples not retrieved. The Odds Ratio (OR) quantifies how enriched the retrieval set is in the target class:

\begin{equation}
    OR = \frac{a \cdot d}{b \cdot c}.
\end{equation}

Previous work has proposed reporting the average OR across queries \citep{MoA_folds_of_enrichment}. However, this approach is unstable when an MoA is associated with very few samples: the denominator terms ($b$ or $c$) can be zero, yielding an infinite OR. Prior works impute these cases with arbitrary, static values, which skews the average even when behavior between methods is otherwise similar \citep{MoA_folds_of_enrichment}. We address this in two ways. First, we apply the modified Haldane-Anscombe correction \citep{modified_haldane_anscombe}, which adds a small constant to all cells of the contingency table, mitigating extreme values without arbitrary imputation. Second, we report the geometric mean rather than the arithmetic mean, following standard statistical practice for log-odds ratios \citep{log_odds_ratio}, which further reduces the influence of outliers.

In addition to the geometric mean OR, we report the percentage of queries achieving statistically significant enrichment. Significance is determined by a permutation test: the observed Odds Ratio ($OR_\text{{obs}}$) is compared against Odds Ratios computed from randomly selected sample sets ($OR_\text{{rand}}$) of the same size. Using $M = 100$ random permutations, the $P$-value for the $j$-th query is:
\begin{equation}
    P_j = \frac{1}{M} \sum_{i=1}^{M} \mathbbm{1}\left(OR_\text{{rand}}^{(i)} \geq OR_\text{{obs}}\right)
\end{equation}
where $\mathbbm{1}(\cdot)$ is the indicator function. The final fraction-significant metric is
\begin{equation}
    P_\text{{sig}} = \frac{\sum_{j=1}^{N} \mathbbm{1}(P_j < 0.05)}{N} \times 100
\end{equation}
where $N$ is the total number of queries. Reporting both metrics is necessary because they capture complementary failure modes: the geometric mean OR can be skewed by a small number of highly enriched queries, while the fraction significant may overly reward methods with low recall but consistent enrichment across many queries.

\subsubsection*{A.3 Gene Pathway Enrichment}
\label{app:pathway}

This task parallels MoA enrichment but uses genetic perturbations, assessing whether the phenotypic profile of a CRISPR knockout retrieves other genes within the same biological pathway or protein complex. Including a distinct perturbation modality allows us to evaluate whether models generalize beyond chemical phenotypes.

We evaluate this task on the \textbf{cpg-CRISPR} subset, which consists of CRISPR-Cas9 knockouts collected from a single institution. Retrieval is evaluated at two stringency levels: No Restriction and NSB. NSS is not applicable as the data come from a single source.

The metric framework is identical to that of MoA enrichment (Section A.2): we report the geometric mean Odds Ratio and the percentage of queries with significant enrichment under permutation testing. The only difference is the source of ground-truth labels, which are derived from five databases of gene pathways and protein complexes (CORUM, HuMAP, StringDB, SIGNOR, Reactome). These databases are described in \hyperref[app:databases]{Appendix B.3.}. Each query gene is evaluated separately against each database, and we report metrics per database.

\subsubsection*{A.4 Replicate Retrieval and Mean Average Precision}
\label{app:discrimination}

\paragraph{kNN Replicate Retrieval.}
This task measures a model's ability to retrieve biological replicates of the same perturbation via $k$-Nearest Neighbor (kNN) search. Unlike the enrichment tasks, replicate retrieval operates on individual well-level profiles rather than consensus representations, directly testing whether a model places replicates of the same perturbation in close proximity in the embedding space.

For each query well, we identify the nearest neighbor by cosine similarity within the candidate set defined by the stringency level. Accuracy is the fraction of queries whose nearest neighbor shares the same perturbation. To account for class imbalance, we use a two-step macro-averaging strategy: accuracy is first calculated per perturbation, then averaged across perturbations to yield the final reported metric.

We evaluate this task on two subsets of the JUMP-CPG dataset:
\begin{itemize}
    \item \textbf{cpg-target2}: A curated subset of 301 compounds with known strong phenotypic effects, replicated across all sources. Evaluated at all four stringency levels.
    \item \textbf{cpg-compound}: A larger, noisier subset spanning three institutions (Sources 5, 9, and 11), testing robustness in a high-throughput, sparse-replicate regime. Evaluated at all four stringency levels. We have downsampled to ensure that every compound has a replicate to match to in every level of stringency.
\end{itemize}

The primary indicator of model robustness is not absolute accuracy at any single level, but the degree of degradation from baseline to the most stringent setting.

\paragraph{Mean Average Precision (mAP)} 
For each query well, we rank all eligible candidate wells by cosine similarity and treat wells with the same perturbation as positives. Average precision summarizes how highly these matching replicates appear in the ranked list, and we average this quantity across queries to obtain the final mAP score. Unlike Recall@1, which only evaluates the nearest neighbor, mAP captures whether same-perturbation replicates are consistently amongst the top of the ranking.

\paragraph{Negative Control Mean Average Precision (NegCon mAP)}
We additionally compute mAP after restricting distractor candidates to DMSO negative-control wells while preserving same-compound replicates as positives. This variant measures whether a representation separates perturbed cells from baseline morphology, rather than requiring it to discriminate every compound from every other compound. As with replicate retrieval, we evaluate each query under NR, NSB, NSS, and NSL scenarios.

\begin{figure}[htbp]
\centering
\qcNegativeControlMapHeatmapBlock{csall-plate-pca64-madctrl-plate-nosph}
\caption{Negative-control mean average precision after 5\% well-level cell-count QC. DMSO wells are used as distractor candidates and same-compound wells are retained as positives. Restrictions: NR No Restriction, NSB = Not Same Batch, NSS = Not Same Source, NSL = Not Same Layout.}
\label{fig:negative-control-map}
\end{figure}

\subsection*{Appendix B: Datasets and Labels}

\subsubsection*{B.1 Dataset Descriptions}

\begin{table}[h] 
    \centering
    \renewcommand{\arraystretch}{1.2} 
    \caption{Summary of the included datasets used for evaluation.}

    \begin{tabular}{lccccccc}
        \toprule
        \textbf{Dataset Name} & \textbf{\makecell{Sources}} & \textbf{\makecell{Batches}} & \textbf{\makecell{Plates}} & 
        \textbf{\makecell{Wells}} &
        \textbf{\makecell{Biological\\Groups}} &
        \textbf{\makecell{Unique\\Perturbations}}\\
        \midrule
        BBBC036 & 1 & 1 & 406 & 153,370 & 676 & 1,551 \\
        cpg-MoA & 10 & 104 & 1,562 & 24,309 & 698 & 1,845 \\
        cpg-CRISPR & 1 & 6 & 148 & 56,832 & 5,287\textsuperscript{$\dagger$} & 7,974 \\
        cpg-target2 & 11 & 107 & 141 & 64,485 & - & 301 \\
        cpg-compound & 3 & 49 & 460 & 157,462 & - & 30,138\\
        \bottomrule
    \end{tabular}
    \label{tab:dataset_details}
    
    \raggedright
    {\footnotesize \textsuperscript{$\dagger$}Mean number of biological groups (protein complexes, clusters, and pathways from CORUM, HuMAP, and Reactome) that contain at least two genes present in our perturbation set. Pairwise databases (StringDB, SIGNOR) are excluded from this count.}
\end{table}

\paragraph{BBBC036.} A single-institution Cell Painting dataset from the Broad Institute, part of the Broad Bioimage Benchmark Collection. It contains chemical perturbations and is used as an upper-bound estimate of MoA retrieval performance in the absence of cross-site batch effects \citep{bbbc036_dataset}.

\paragraph{cpg-MoA.} A subset of the JUMP-CPG dataset containing chemical perturbations with mechanism-of-action annotations transferred from the Drug Repurposing Hub. Compounds are profiled across multiple experimental batches and institutions, enabling evaluation of MoA retrieval under increasing batch-effect stringency. This dataset complements BBBC036 by testing whether representations recover shared mechanisms of action in a larger, multi-source Cell Painting setting \citep{cpg0016_dataset, drug_repurposing_hub}.
 
\paragraph{cpg-CRISPR.} A subset of the JUMP-CPG dataset containing CRISPR-Cas9 genetic knockouts from a single institution. This dataset enables evaluation of gene pathway enrichment and provides a distinct perturbation modality from the chemical compound datasets \citep{cpg0016_dataset}.

\paragraph{cpg-target2.} A subset of the JUMP-Cell Painting (JUMP-CPG) dataset containing approximately 300 chemical compounds curated for known strong phenotypic effects. Compounds are replicated across multiple institutions, enabling evaluation at all four stringency levels \citep{cpg0016_dataset}.

\paragraph{cpg-compound.} A larger subset of the JUMP-CPG dataset containing chemical perturbations across three institutions (Sources 5, 9, and 11). Compared to cpg-target2, this dataset includes a much broader range of compounds, many of which produce weak or undetectable morphological effects, making it a more challenging high-throughput screening regime \citep{cpg0016_dataset}.

\subsubsection*{B.2 Mechanism of Action Labels}

MoA labels for compounds in BBBC036 and cpg-MoA are derived from the Drug Repurposing Hub \citep{drug_repurposing_hub}, a curated resource that annotates compounds with their known molecular targets and mechanisms of action. We use these labels to define ground-truth class membership for the MoA enrichment task: two compounds are considered to share an MoA if they are annotated with the same mechanism in the Drug Repurposing Hub.

\subsubsection*{B.3 Gene Pathway and Protein Complex Databases}
\label{app:databases}

For the gene pathway enrichment task, ground-truth gene pairs are derived from five complementary databases. Each database captures a different type of functional relationship, and we report metrics per database to surface differences in how methods generalize across relationship types.

\begin{itemize}
    \item \textbf{CORUM} \citep{CORUM}: A manually curated repository of experimentally verified mammalian protein complexes.
    \item \textbf{HuMAP} \citep{humap}: A census of human protein complexes derived from integration of mass spectrometry experiments.
    \item \textbf{StringDB} \citep{StringDB}: A database of known and predicted protein-protein interactions, encompassing both physical associations and functional linkages. This is the largest database in our evaluation.
    \item \textbf{SIGNOR} \citep{SIGNOR}: A signaling network database. This is the smallest database in our evaluation, which limits statistical power for enrichment testing.
    \item \textbf{Reactome} \citep{Reactome}: A curated knowledgebase of biological pathways and molecular reactions in human biology.
\end{itemize}

\subsection*{Appendix C: Models and Pipelines}

\subsubsection*{C.1 Models Evaluated}
\label{app:models}

We curate a cross-section of representation extraction strategies currently employed in morphological profiling, spanning classical feature engineering, transfer learning from natural-image foundation models, and microscopy-specific pretrained models.

\paragraph{CellProfiler.} The industry standard for over a decade, CellProfiler serves as our primary non-deep-learning baseline \citep{carpenter_cellprofiler}. The pipeline relies on classical image processing to segment cells and measure predefined, interpretable morphological features including cell size, shape, texture, and intensity distribution. It represents the performance of expert-defined, hand-crafted feature engineering. We used the CellProfiler features released with the BBBC036 \citep{bbbc036_dataset} and the JUMP-CPG \citep{cpg0016_dataset} dataset.

\paragraph{ImageNet ResNet101.} A ResNet101 pretrained on ImageNet in a supervised manner \citep{resnet}. This model serves as a transfer learning baseline, testing whether features learned from natural images are sufficiently general to capture relevant signal in microscopy data without domain-specific fine-tuning. To generate a well-level representation, we average field-of-view (FOV) profiles within each channel and concatenate the five channel-level representations for each well. This yields a 2,048-dimensional embedding per channel and a 10,240-dimensional embedding per well.

\paragraph{Untrained ResNet101.} A ResNet101 architecture with randomly initialized weights, included as a negative control. This baseline quantifies how much performance derives solely from the inductive bias of the convolutional architecture itself, independent of any learned features. We follow the same inference procedure described for ImageNet ResNet101.

\paragraph{DINOv2.} A Vision Transformer trained with self-supervised learning on a large corpus of natural images \citep{dinov2}. DINOv2 leverages a distillation-based objective and the attention mechanism, which often yields more robust semantic representations than supervised CNNs. Because it remains unsettled whether the CLS token embedding or mean-pooled patch embeddings is the better practice for downstream tasks, we evaluate both variants. We use the DINOv2 ViT-B/14 model and follow the same well-level aggregation procedure described for ImageNet ResNet101.

\paragraph{OpenPhenom.} A microscopy-specific masked autoencoder pretrained on large-scale Cell Painting data \citep{OpenPhenom}. OpenPhenom is channel-agnostic, allowing it to handle the multi-channel structure of Cell Painting natively. Including this model lets us measure the specific value of microscopy pretraining over transfer learning from natural images. 

\paragraph{SubCell.} A model pretrained on fluorescent microscopy datasets distinct from Cell Painting, previously demonstrated to transfer well to Cell Painting tasks \citep{subcell}. For SubCell, we follow the feature extraction procedure recommended by \citet{subcell} for applying the model to Cell Painting data.

\paragraph{CLOOME.} A multimodal model trained on Cell Painting images jointly with chemical compound structures \citep{cloome}. This model tests whether incorporating chemical structure as an auxiliary modality during pretraining yields better representations for downstream morphological profiling.

\begin{table}[H]
    \centering
    \caption{Summary of the models evaluated in the benchmark.}
    \label{tab:models}
    \renewcommand{\arraystretch}{1.2}
    \begin{tabular}{lccc}
        \toprule
        \textbf{Model} & \textbf{Pretrained} & \textbf{\makecell{Pretrained on\\Microscopy Images}} & \textbf{Extraction Method} \\
        \midrule
        CellProfiler & \ding{55} & \ding{55} & Classical / Hand-crafted \\
        ResNet101 (ImageNet) & \ding{51} & \ding{55} & Supervised CNN \\
        ResNet101 (Randomized) & \ding{55} & \ding{55} & Untrained CNN \\
        DINOv2 & \ding{51} & \ding{55} & Self-Supervised ViT \\
        OpenPhenom & \ding{51} & \ding{51} & Masked Autoencoder (ViT) \\
        SubCell & \ding{51} & \ding{51} & Self-Supervised ViT \\
        CLOOME & \ding{51} & \ding{51} & Multimodal Contrastive \\
        \bottomrule
    \end{tabular}
\end{table}

\subsubsection*{C.2 Image Preprocessing}

Prior to feature extraction, raw microscopy images were converted to 8-bit PNG format to standardize dynamic range and reduce storage requirements. Illumination correction was then applied to all fields of view to rectify uneven lighting and vignetting artifacts introduced by the optical system.

\subsubsection*{C.3 Sample Quality Control}

We applied a series of filtering steps to remove noisy data at the batch, channel, and well levels.

\paragraph{Batch Filtering.} To ensure experimental consistency, we evaluated batches based on the stability of their negative controls. For each batch, we computed the mean cosine distance of all wells to their nearest negative control using CellProfiler features. Batches where this metric deviated by more than 0.2 from the expected baseline were discarded. The threshold was chosen based on manual verification, which indicated that batches exceeding this value showed extensive contamination or microscopy imaging artifacts.

\paragraph{Channel De-duplication.} Cell Painting collects five fluorescent channels per microscopy image, each capturing a distinct set of cellular compartments. Images with duplicated channels were removed as artifacts.

\paragraph{Cell Count Filtering.} We removed wells with insufficient cell counts. The global cell count distribution was estimated by sampling 5,000 images. Wells falling in the bottom 5th percentile of this distribution were removed.

\subsubsection*{C.4 Feature Normalization Pipeline}

While some prior benchmarks use representations as-is \citep{MoA_folds_of_enrichment}, this approach does not control for varying dimensionality or feature redundancy across methods. In particular, it disadvantages CellProfiler, which produces a high-dimensional feature space that is partially redundant. To ensure fairness across methods, we standardize all representations to the same dimensionality and apply a consistent batch-effect correction pipeline:

\begin{enumerate}
    \item \textbf{Platewise Center Scaling and PCA.} All features are scaled to a standard range, then reduced via Principal Component Analysis applied across the entire dataset. This step both standardizes dimensionality and denoises the feature space. We use 64 components for the PCA analysis.
    \item \textbf{Platewise Robust Standardization.} A Median Absolute Deviation (MAD) "Robustize" transformation is applied per plate to align distributions across experimental runs while remaining robust to outliers.
\end{enumerate}

This pipeline is applied identically to all methods evaluated in the benchmark.

\newpage
\subsection*{Appendix D: Supplementary Analyses}

\subsubsection*{D.1 Effect of Modified Haldane-Anscombe Correction on OR Distribution}
\label{app:haldane}

\begin{figure}[h]
    \centering
    \includegraphics[width=0.8\textwidth]{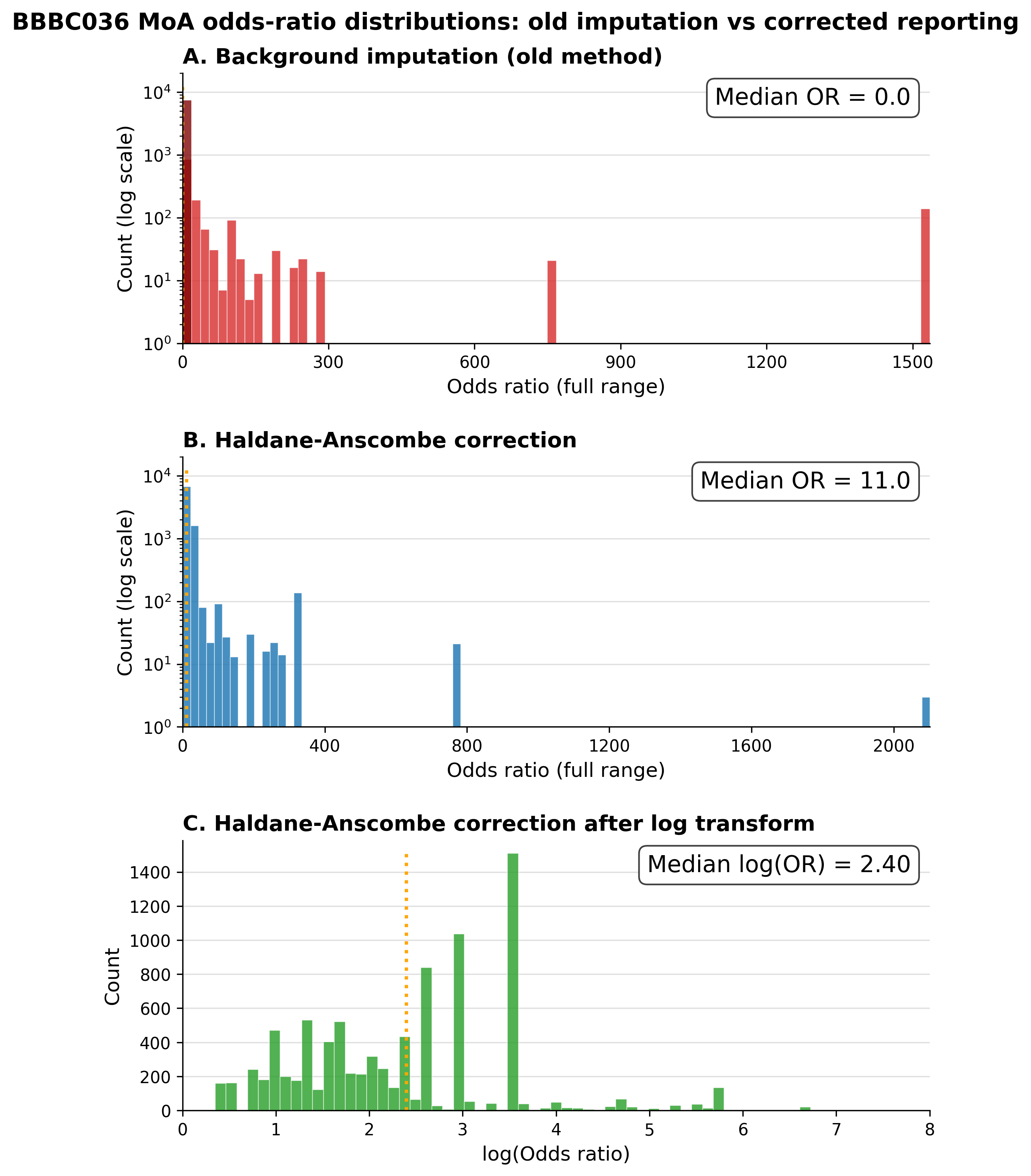}
    \caption{Odds-ratio (OR) distribution for the BBBC036 MoA task (all models pooled, 1\% cutoff) with the \textbf{(A)} Old imputation method, \textbf{(B)} modified Haldane-Anscombe (mHA) correction, and
    \textbf{(C)} mHA correction and log-transformed OR.}
    \label{fig:or_imputation}
\end{figure}

When a biological class has few members, all members can appear in the top-$k$ retrieval set, making the OR undefined. \citet{MoA_folds_of_enrichment} handled this by imputing infinite values with the total size of the background set. Because the size of the background set is very large, this creates extreme outliers that dominate the arithmetic mean. In the BBBC036 MoA task, 83\% of per-query OR values are exactly zero, yet the arithmetic mean is 32.1 (Appendix Figure~\ref{fig:or_imputation}, Panel A), making it unrepresentative of typical behavior.

We replace this with the modified Haldane-Anscombe (mHA) correction \citep{modified_haldane_anscombe}, which adds $\frac{1}{2}$ to all four cells of the contingency table only when at least one cell is zero. We use the modified variant rather than the original Haldane-Anscombe correction, which adds $\frac{1}{2}$ unconditionally. Applying the correction to all tables biases the expected OR away from 1 under the null, because the additive constant is non-negligible relative to cell counts in small tables, shifting the baseline.

Even after mHA correction, the distribution remains heavily right-skewed. Following standard practice for ratio-scale effect sizes \citep{log_odds_ratio}, we report the geometric mean, which compresses the right tail and yields a nearly symmetric distribution.

\newpage
\subsubsection*{D.2 Gene Pathway Enrichment per Database}
\label{app:gene_path}

\begin{figure}[h]
    \centering
    \includegraphics[width=\textwidth]{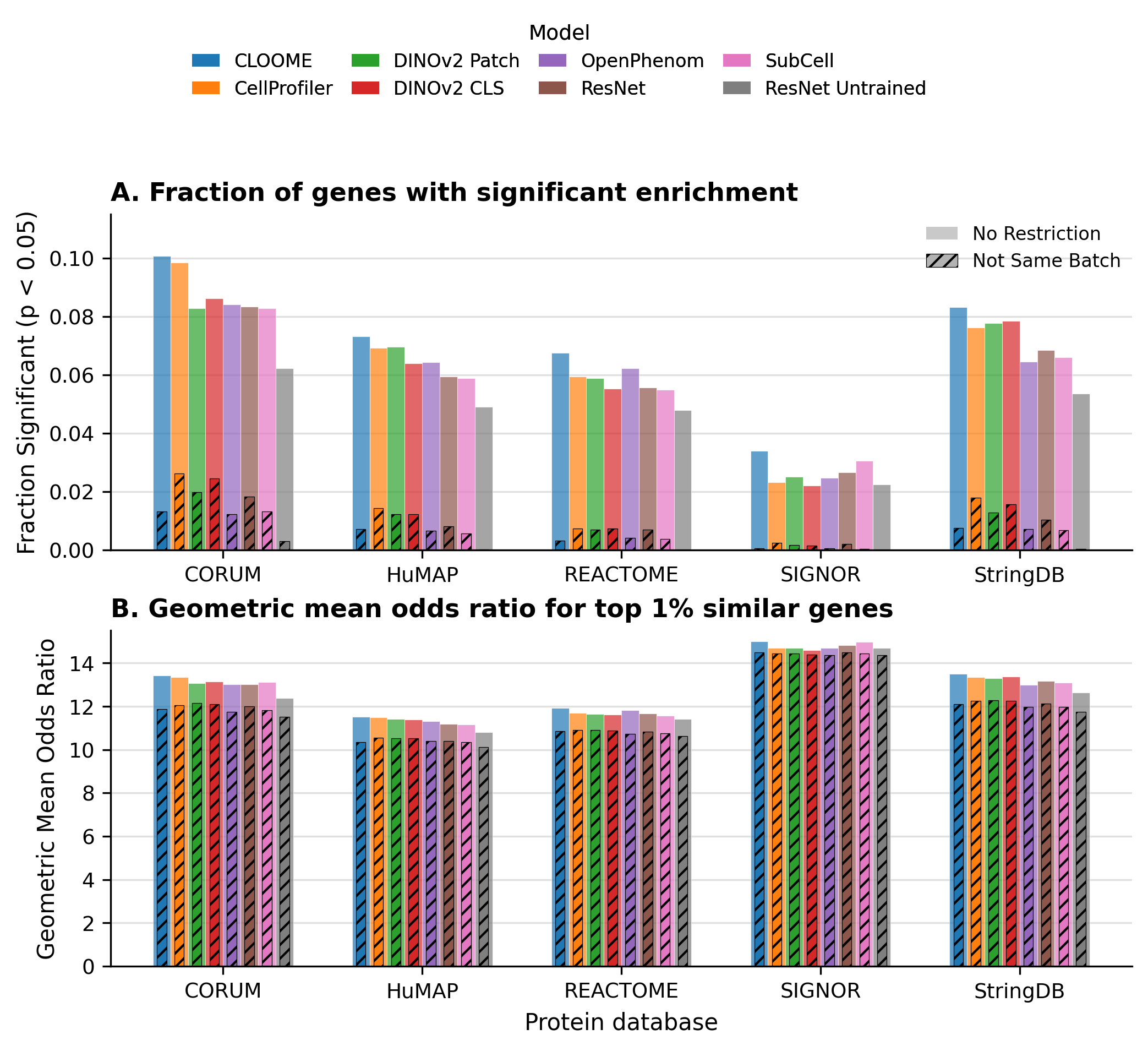}
    \caption{Gene pathway enrichment performance by database. Bars show model
    performance separately for each gene-set database. Transparent bars indicate No
    Restriction results, while hatched bars indicate Not Same Batch results. Panel A
    reports the fraction of genes with significant enrichment, and Panel B reports the geometric mean odds ratio.}
    \label{fig:gene_pathway_per_db}
\end{figure}

\newpage
\subsubsection*{D.3 Brittleness of Clustering Metrics under Random Label Permutation}
\label{app:clustering}
\begin{table}[h]
    \centering
    \caption{Raw scores and performance gap ($\Delta$ = scVI $-$ Baseline) under original labels and three random label permutations, using the evaluation framework of \citet{arevalo_batch_effects}. The $\Delta$ in overall score persists under shuffling, indicating it reflects latent space structure rather than biological signal.}
    \label{tab:arevalo_shuffle}
    \renewcommand{\arraystretch}{1.15}
    \resizebox{\textwidth}{!}{%
    \begin{tabular}{llcccccccccccccc}
        \toprule
        & & \multicolumn{4}{c}{\textbf{Batch Correction}} & \multicolumn{6}{c}{\textbf{Bio Metrics}} & \multicolumn{3}{c}{\textbf{Aggregate}} \\
        \cmidrule(lr){3-6} \cmidrule(lr){7-12} \cmidrule(lr){13-15}
        \textbf{Condition} & \textbf{Method} & \makecell{Graph\\Conn.} & KBET & \makecell{LISI\\batch} & \makecell{Silh.\\batch} & \makecell{LISI\\label} & \makecell{Leiden\\ARI} & \makecell{Leiden\\NMI} & \makecell{Silh.\\label} & \makecell{mAP\\(ctrl)} & \makecell{mAP\\(nonrep)} & \makecell{Batch\\Corr.} & Bio & \textbf{Overall} \\
        \midrule
        \multirow{3}{*}{\makecell[l]{Original\\labels}} 
        & scVI     & 0.34 & 0.04 & 0.25 & 0.78 & 1.00 & 0.02 & 0.28 & 0.25 & 0.25 & 0.06 & 0.35 & 0.31 & 0.33 \\
        & Baseline & 0.34 & 0.01 & 0.01 & 0.80 & 1.00 & 0.00 & 0.23 & 0.31 & 0.21 & 0.03 & 0.29 & 0.30 & 0.29 \\
        \cdashline{2-15}
        & $\Delta$ & 0.00 & \textbf{+.03} & \textbf{+.24} & $-$.02 & .00 & +.02 & +.05 & $-$.06 & +.04 & +.03 & \textbf{+.06} & +.01 & \textbf{+.04} \\
        \midrule
        \multirow{3}{*}{\makecell[l]{Shuffle\\seed 1}} 
        & scVI     & 0.32 & 0.00 & 0.36 & 0.88 & 1.00 & 0.00 & 0.22 & 0.19 & 0.16 & 0.02 & 0.39 & 0.27 & 0.32 \\
        & Baseline & 0.32 & 0.00 & 0.03 & 0.89 & 1.00 & 0.00 & 0.22 & 0.25 & 0.14 & 0.01 & 0.31 & 0.27 & 0.29 \\
        \cdashline{2-15}
        & $\Delta$ & 0.00 & .00 & \textbf{+.33} & $-$.01 & .00 & .00 & .00 & $-$.06 & +.02 & +.01 & \textbf{+.08} & .00 & \textbf{+.03} \\
        \midrule
        \multirow{3}{*}{\makecell[l]{Shuffle\\seed 2}} 
        & scVI     & 0.32 & 0.00 & 0.35 & 0.87 & 1.00 & 0.00 & 0.22 & 0.19 & 0.17 & 0.00 & 0.39 & 0.26 & 0.31 \\
        & Baseline & 0.32 & 0.00 & 0.03 & 0.89 & 1.00 & 0.00 & 0.22 & 0.25 & 0.14 & 0.01 & 0.31 & 0.27 & 0.29 \\
        \cdashline{2-15}
        & $\Delta$ & 0.00 & .00 & \textbf{+.32} & $-$.02 & .00 & .00 & .00 & $-$.06 & +.03 & $-$.01 & \textbf{+.08} & $-$.01 & \textbf{+.02} \\
        \midrule
        \multirow{3}{*}{\makecell[l]{Shuffle\\seed 3}} 
        & scVI     & 0.32 & 0.00 & 0.35 & 0.88 & 1.00 & 0.00 & 0.23 & 0.19 & 0.16 & 0.02 & 0.39 & 0.27 & 0.31 \\
        & Baseline & 0.32 & 0.00 & 0.03 & 0.89 & 1.00 & 0.00 & 0.22 & 0.25 & 0.14 & 0.01 & 0.31 & 0.27 & 0.29 \\
        \cdashline{2-15}
        & $\Delta$ & 0.00 & .00 & \textbf{+.32} & $-$.01 & .00 & .00 & +.01 & $-$.06 & +.02 & +.01 & \textbf{+.08} & .00 & \textbf{+.02} \\
        \bottomrule
    \end{tabular}%
    }
\end{table}

\subsubsection*{D.4 Postprocessing Fairness}
\label{app:cellprofiler-dim}

The models compared in this benchmark produce feature spaces with very different dimensionalities and structures. Classical CellProfiler features are high-dimensional and contain many correlated measurements of related cellular properties; learned embeddings are typically lower-dimensional, compressed representations whose internal scaling is determined by the training objective. In order to ensure a degree of fairness in evaluation across models we have chosen a minimal postprocessing pipeline. Otherwise, there is a risk of measuring differences in feature geometry as much as differences in biological content. However, we note that the postprocessing steps do not have neutral effect across all models.

The analyses below show that the model rankings are not invariant to these choices. Applying plate-wise CenterScale normalization, PCA, and robust batch-wise normalization produces a setting in which CellProfiler is particularly strong for replicate retrieval. This is biologically plausible: many CellProfiler measurements are redundant or highly correlated, and dimensionality reduction can concentrate a shared morphological signal while suppressing unstable axes. 

When PCA is removed, learned embeddings become substantially more competitive, and in several panels dominate the ranking. This does not imply that learned models are intrinsically superior, nor that CellProfiler is intrinsically weak. Rather, it suggests that high-dimensional handcrafted features can be penalized when evaluated in their raw embedding, where batch effects and uneven feature scales may occupy many dimensions. Learned embeddings, by contrast, have already undergone a form of compression during model training and may therefore be less dependent on an external dimensionality-reduction step. It is also important to note that ResNet benefits substantially from PCA, which is expected because each well is represented by a high dimensional embedding: 2,048 features per channel, or 10,240 features per well after concatenating all five channels.

CenterScale normalization also benefits CellProfiler the most. This is expected
because classical CellProfiler features span a wide range of scales, which can cause some measurements to dominate the embedding space without normalization.

Batch-wise sphering shows a more mixed pattern. Whitening can reduce batch-specific covariance structure and may help enrichment tasks that rely on consensus profiles, where averaging already suppresses some well-level noise. However, the same operation can be detrimental for nearest-neighbor replicate retrieval, where individual-well geometry is the object of evaluation. In that setting, sphering may overcorrect batch structure that is partially entangled with biological signal. This illustrates why a correction that is reasonable for one biological retrieval task may not be uniformly beneficial across all tasks.

For these reasons, we treat the main postprocessing pipeline as minimal as possible while ensuring a fair comparison. The panels below show how results would change under plausible alternatives. The important conclusion is that postprocessing can shape benchmark rankings and should be reported as part of the benchmark specification.

\qcPipelineMoaFigure{csall-plate-pca64-madctrl-plate-sphctrl-batch}
{batch-sphered}
{qc-sphctrl-batch}

\qcPipelineCrisprFigure{csall-plate-pca64-madctrl-plate-sphctrl-batch}
{batch-sphered}
{qc-sphctrl-batch}

\qcPipelineKnnFigure{csall-plate-pca64-madctrl-plate-sphctrl-batch}
{batch-sphered}
{qc-sphctrl-batch}

\qcPipelineMoaFigure{nocs-pca64-madctrl-plate-nosph}
{no center scaling}
{qc-nocs-pca64-nosph}

\qcPipelineCrisprFigure{nocs-pca64-madctrl-plate-nosph}
{no center scaling}
{qc-nocs-pca64-nosph}

\qcPipelineKnnFigure{nocs-pca64-madctrl-plate-nosph}
{no center scaling}
{qc-nocs-pca64-nosph}

\qcPipelineMoaFigure{csall-plate-nopca-madctrl-plate-nosph}
{no PCA}
{qc-nopca-nosph}

\qcPipelineCrisprFigure{csall-plate-nopca-madctrl-plate-nosph}
{no PCA}
{qc-nopca-nosph}

\qcPipelineKnnFigure{csall-plate-nopca-madctrl-plate-nosph}
{no PCA}
{qc-nopca-nosph}

\newpage
\clearpage

\begin{figure}[h]
\centering
\includegraphics[width=\textwidth]{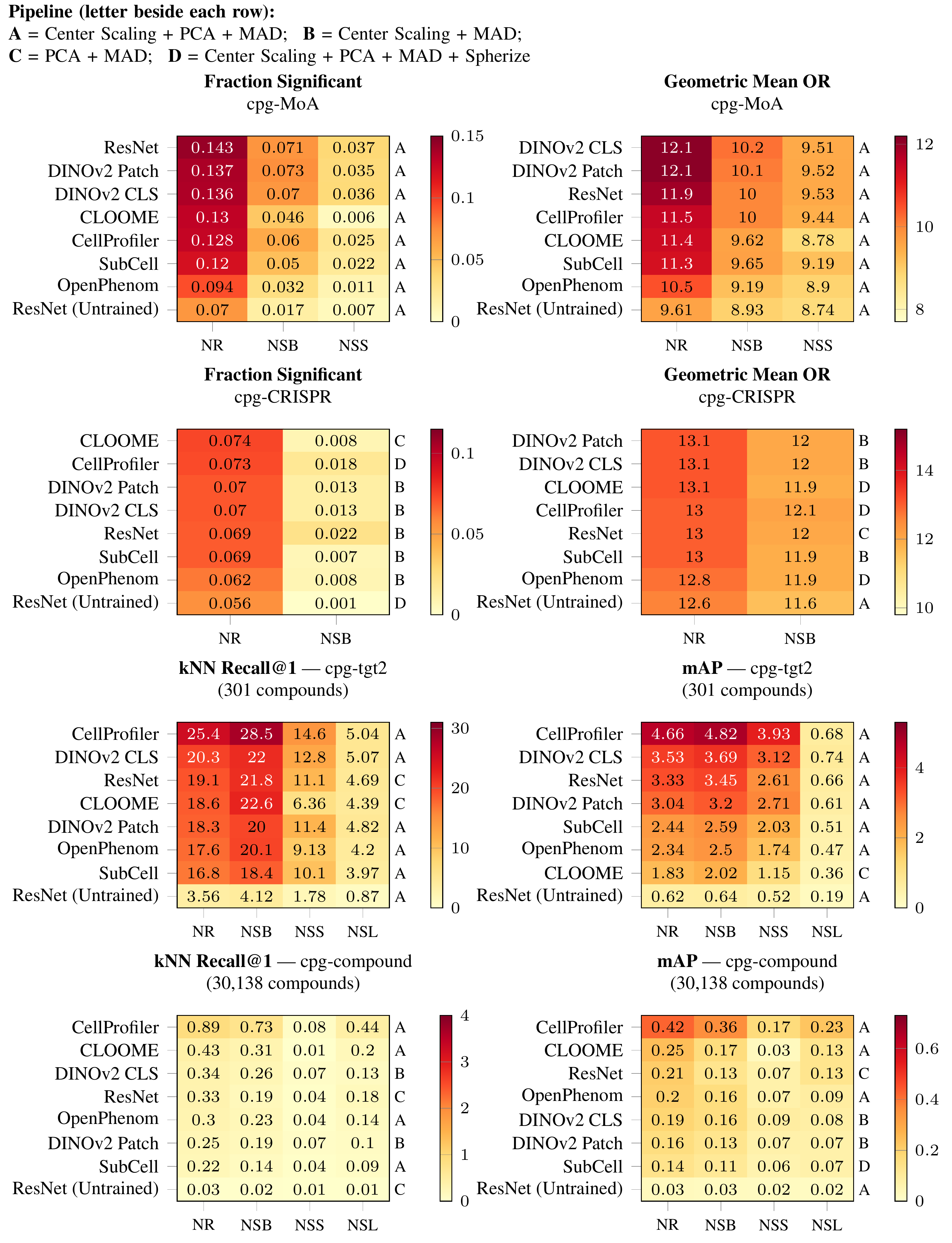}
\caption{Best performance of each model across postprocessing pipelines. For each dataset, metric, and stringency level, the highest score achieved by each model across all postprocessing variants in Appendix D.4. Labels to the right of each row give the pipeline that achieved it.}
\label{fig:best-postprocessing}
\end{figure}

To assess whether our conclusions depend on a single shared pipeline, we also report the best performance of each model across all postprocessing variants below (Appendix Figure~\ref{fig:best-postprocessing}).

\newpage

\subsubsection*{D.5 Global versus Unleaked PCA}
\label{app:unleaked-pca}
Our main pipeline fits PCA on the entire dataset, including the samples used as queries and retrieval targets. Whether this is appropriate depends on the benchmark. When the aim is to integrate an existing multi-institution dataset, such as the cpg datasets (cpg-moa, cpg-compound) where the data are freely exchanged across institutions but carry technical differences that make naive comparison impossible. In this scenario, a global PCA is appropriate: the benchmark does not assess generalization to unseen data, but whether representations sufficiently minimize spurious technical variation. Here we also provide where the PCA is only fit based on the task (e.g. fit on one source for NSS and compared against the rest of the sources).

\begin{figure}[h]
\centering
\includegraphics[width=\textwidth]{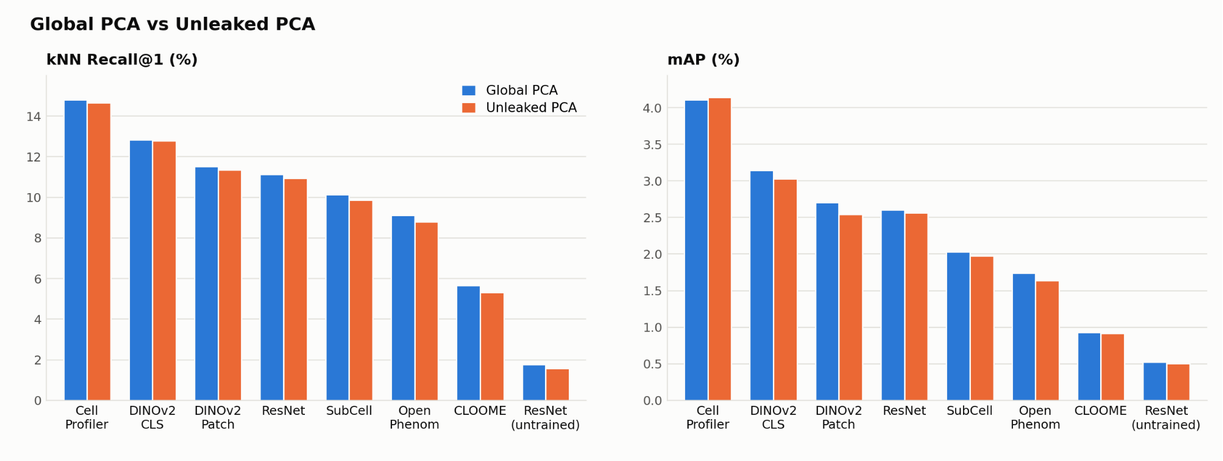}
\caption{Global versus unleaked PCA. Compound replicate retrieval on cpg-tgt2 in the Not in Same Source setting, with PCA fit on the entire dataset (Global PCA) or only on the query source (Unleaked PCA). \textbf{Left:} kNN Recall@1 (\%). \textbf{Right:} mAP (\%).}
\label{fig:unleaked-pca}
\end{figure}

\newpage
\subsubsection*{D.6 F1 and Recall for Replicate Retrieval}
\label{app:f1}

In addition to kNN accuracy and mAP, we evaluate replicate retrieval with the F1 score and recall.

\begin{figure}[h]
\centering
\includegraphics[width=\textwidth]{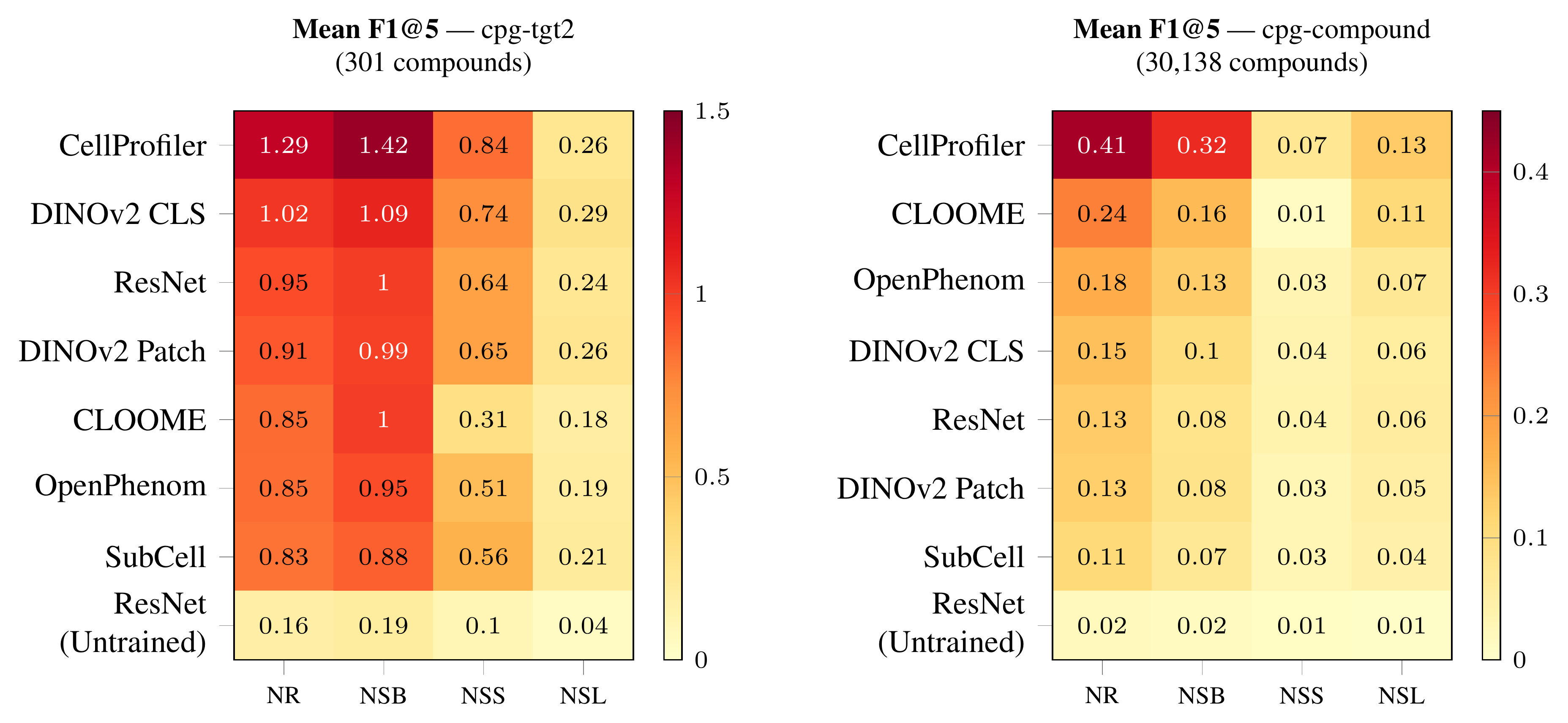}
\caption{Mean F1@5 for kNN replicate retrieval. Mean F1@5 (\%) of each model on cpg-tgt2 (left) and cpg-compound (right) across stringency levels.}
\label{fig:f1-knn}
\end{figure}

\begin{figure}[h]
\centering
\includegraphics[width=0.6\textwidth]{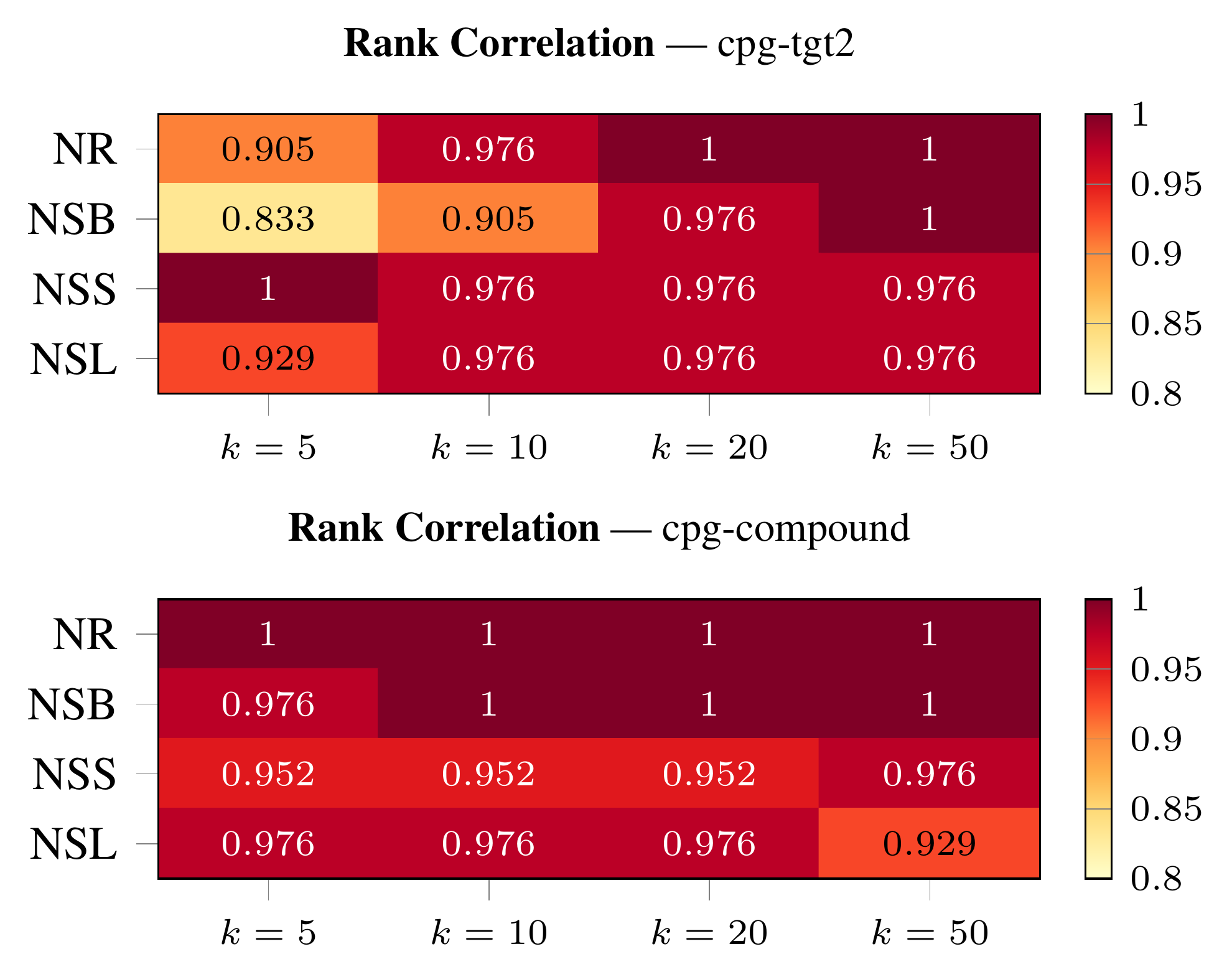}
\caption{Agreement between mAP and F1 rankings. Spearman rank correlation between model rankings under mAP and under mean F1@$k$, for each stringency level and $k$, on cpg-tgt2 (top) and cpg-compound (bottom).}
\label{fig:f1-rank-corr}
\end{figure}

\newpage
\clearpage
\subsubsection*{D.7 Statistical Robustness of Model Rankings}
\label{app:stats}

\begin{figure}[H]
\centering
\includegraphics[width=\textwidth]{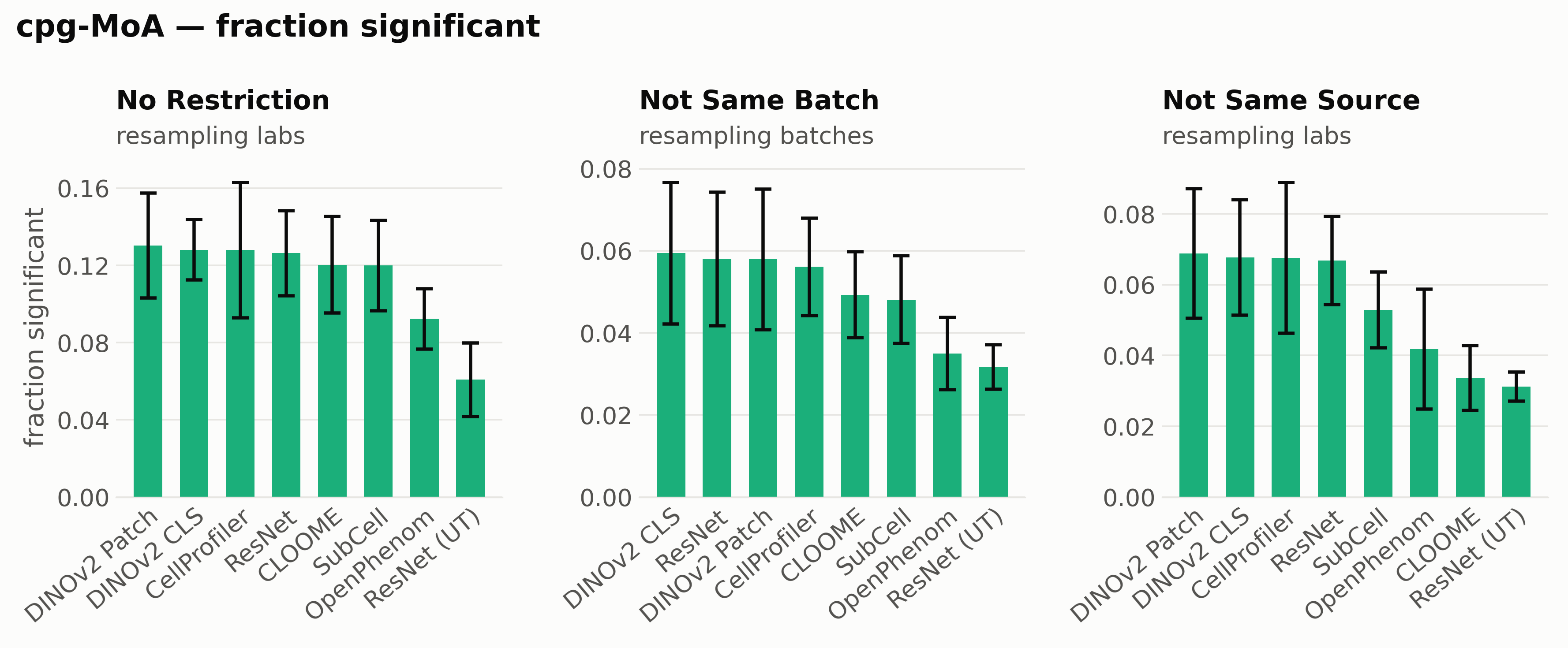}
\caption{MoA enrichment fraction significant under resampling.}
\label{fig:boot-moa-fs}
\end{figure}

\begin{figure}[H]
\centering
\includegraphics[width=\textwidth]{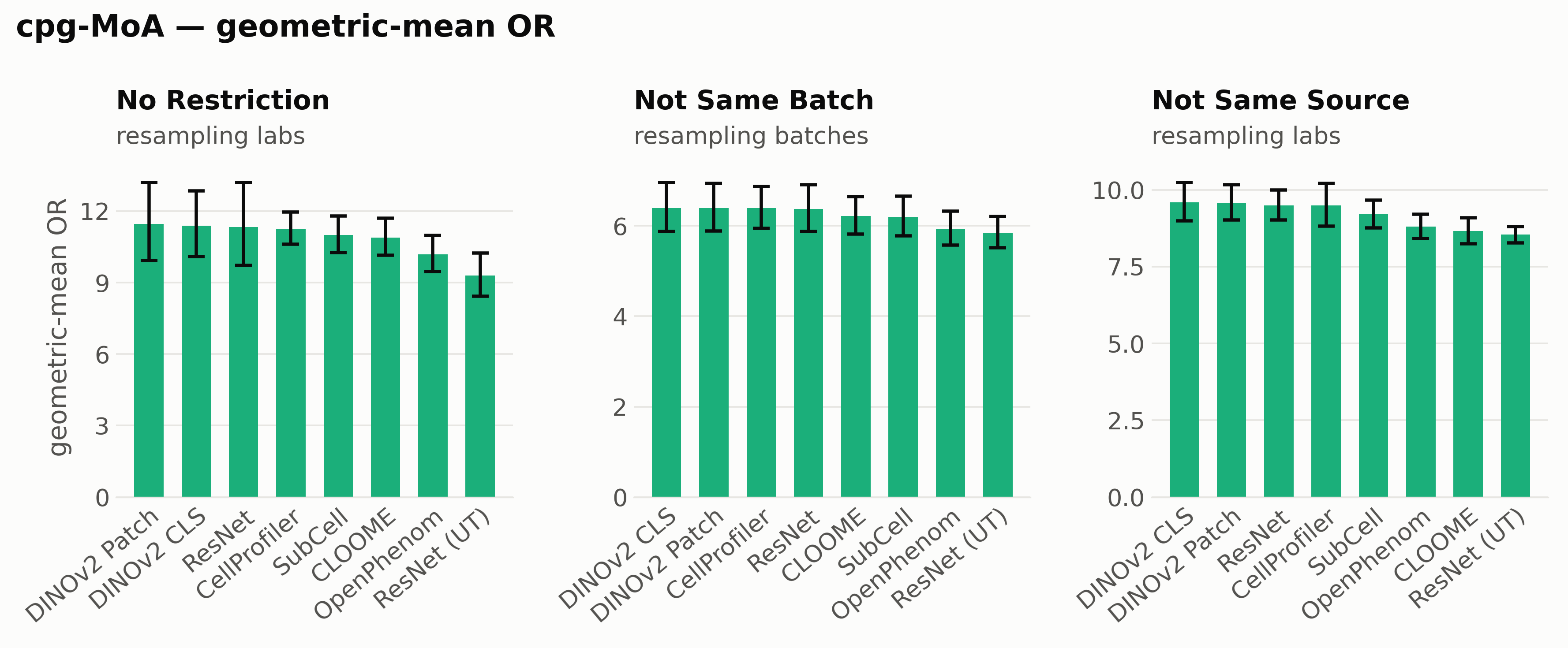}
\caption{MoA enrichment geometric mean OR under resampling.}
\label{fig:boot-moa-or}
\end{figure}

\begin{figure}[H]
\centering
\includegraphics[width=0.684\textwidth]{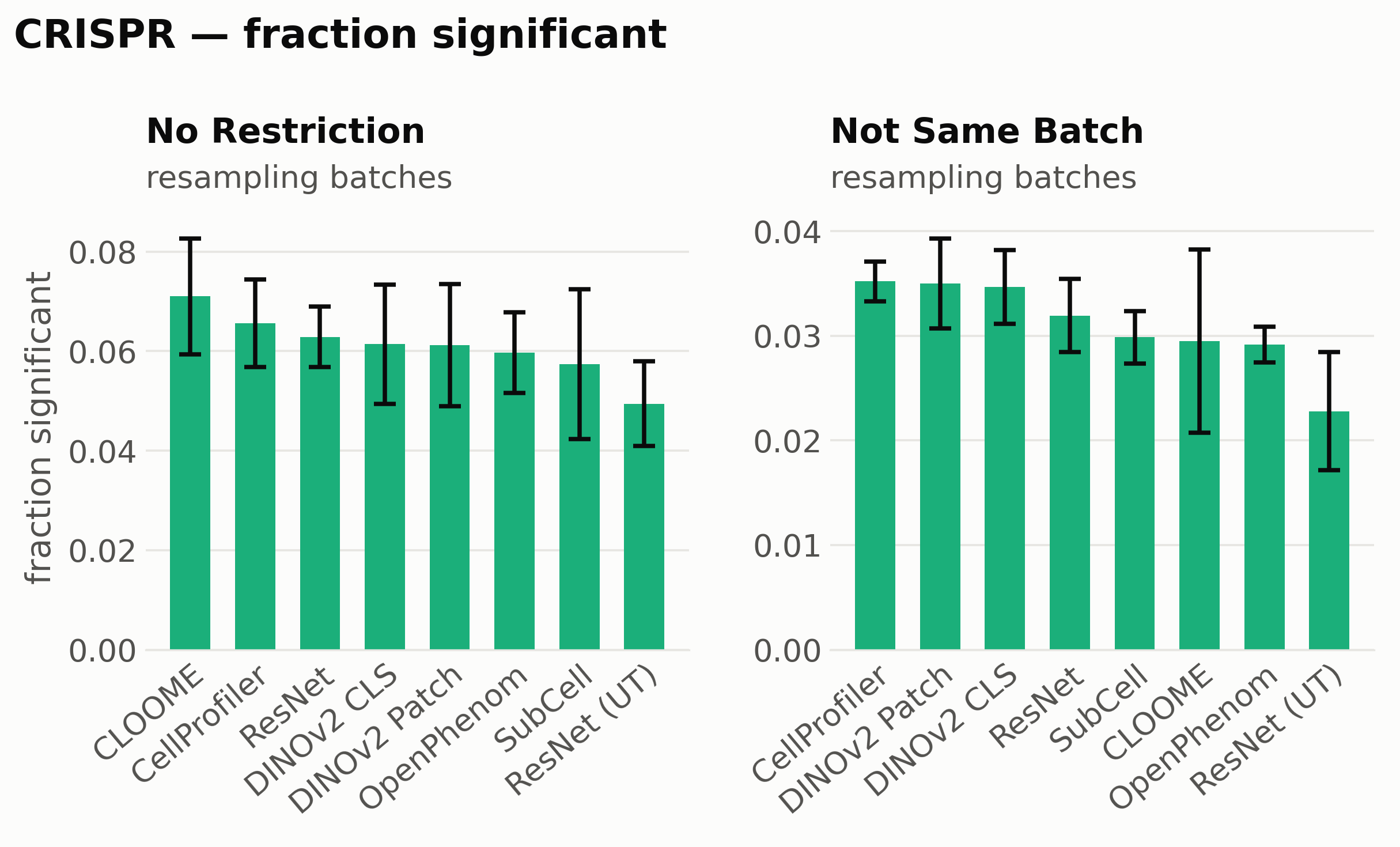}
\caption{Gene pathway enrichment fraction significant under resampling.}
\label{fig:boot-crispr-fs}
\end{figure}

\begin{figure}[H]
\centering
\includegraphics[width=0.684\textwidth]{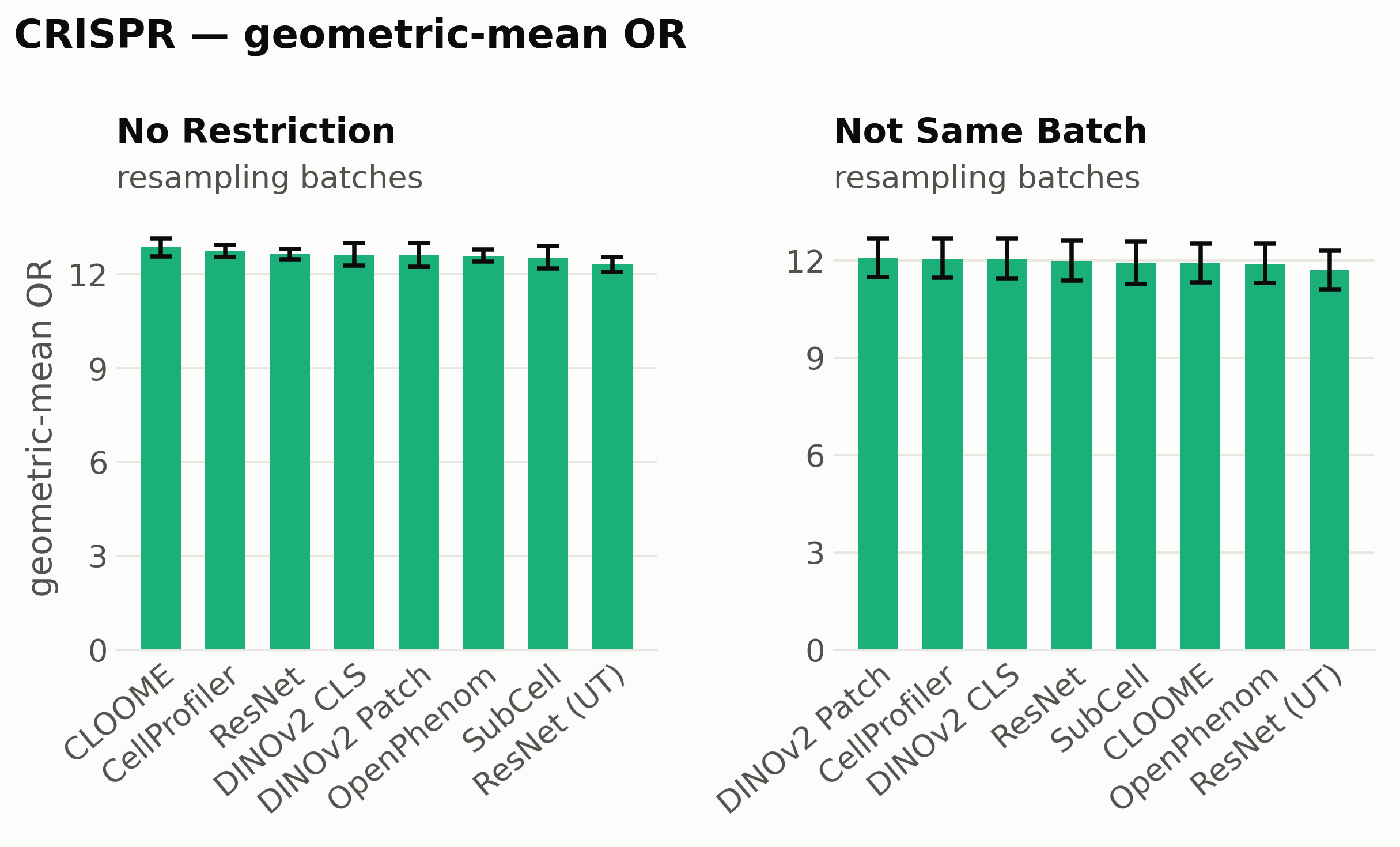}
\caption{Gene pathway enrichment geometric mean OR under resampling.}
\label{fig:boot-crispr-or}
\end{figure}

\begin{figure}[H]
\centering
\includegraphics[width=\textwidth]{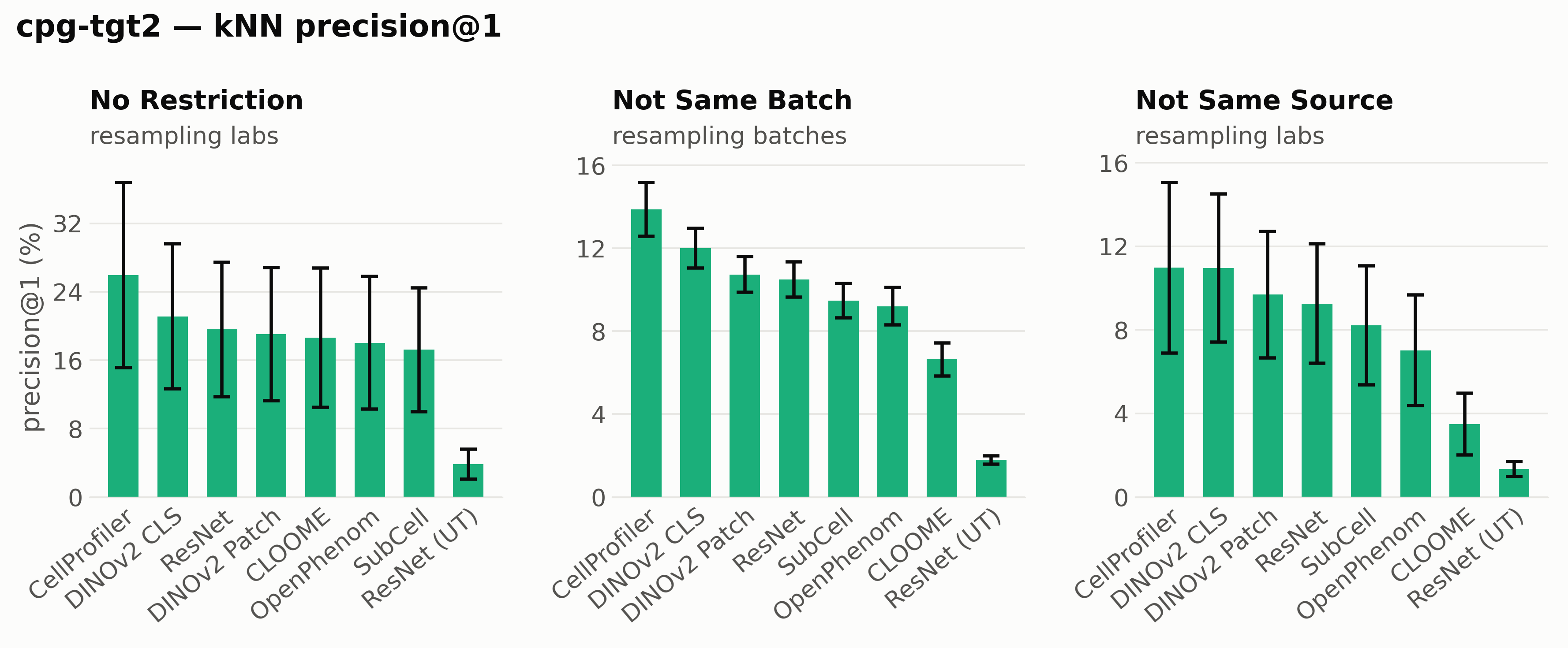}
\caption{Replicate retrieval kNN precision@1 under resampling.}
\label{fig:boot-tgt2-p1}
\end{figure}

\begin{figure}[H]
\centering
\includegraphics[width=\textwidth]{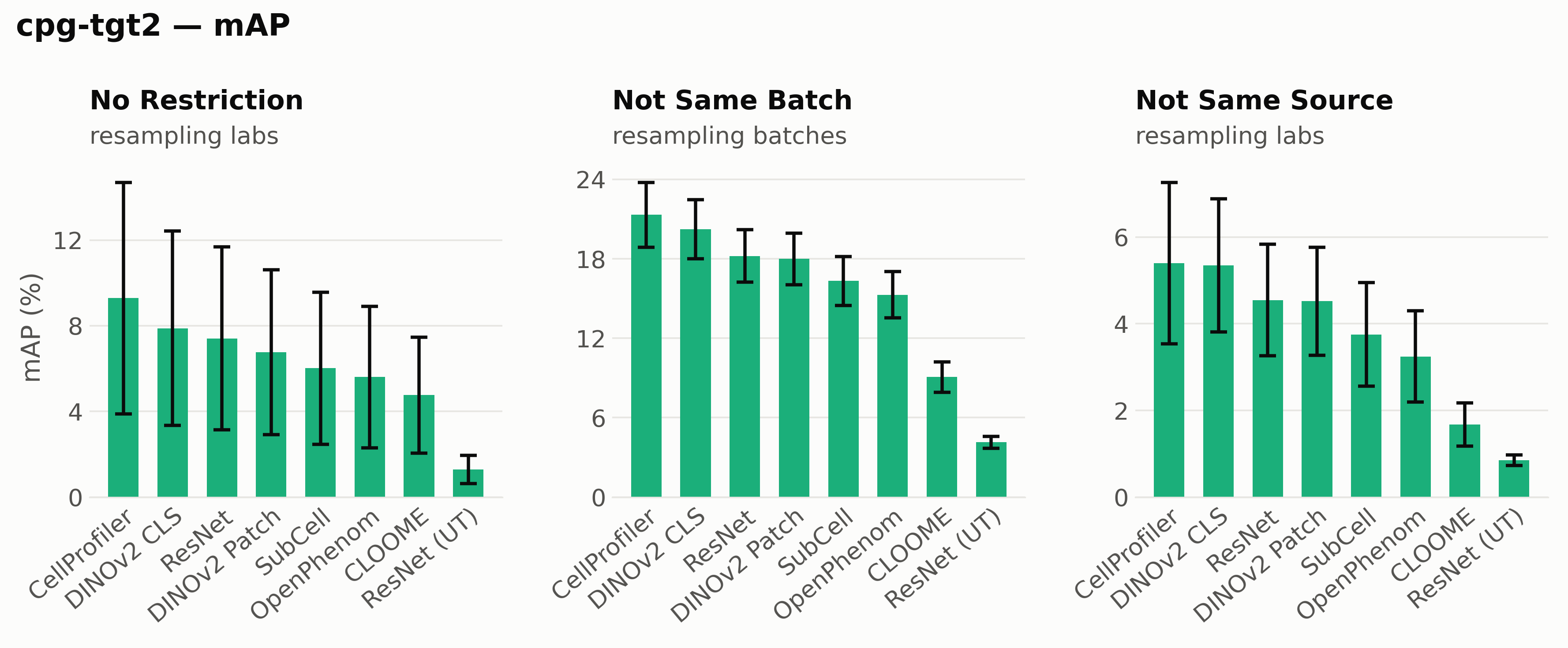}
\caption{Replicate retrieval mAP under resampling.}
\label{fig:boot-tgt2-map}
\end{figure}

\begin{figure}[H]
\centering
\includegraphics[width=\textwidth]{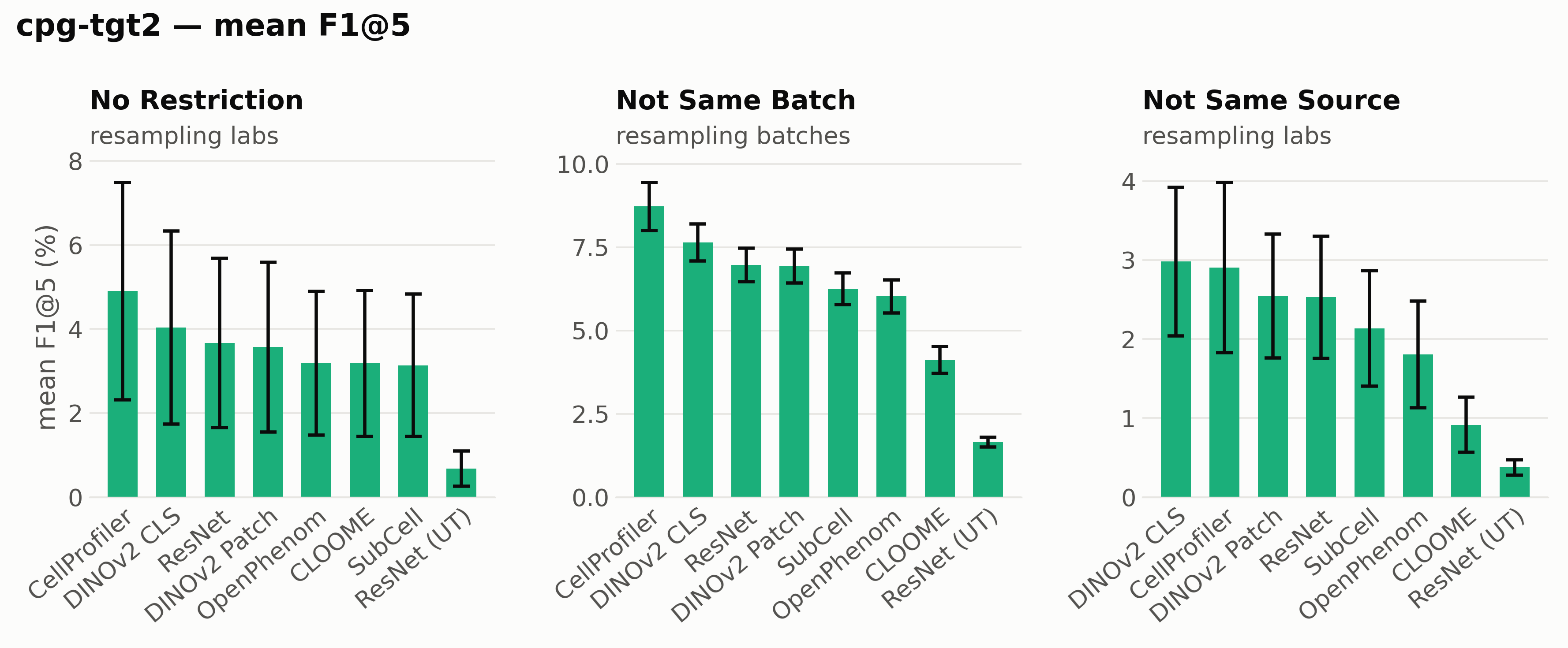}
\caption{Replicate retrieval mean F1@5 under resampling.}
\label{fig:boot-tgt2-f1}
\end{figure}

\paragraph{Resampling sources and batches.} To assess whether model rankings hold across laboratories and experimental batches, we bootstrap the sources of technical variation underlying each setting including: (1) sources in the No Restriction and Not in Same Source settings, and (2) batches in the Not in Same Batch setting and for the single-source cpg-CRISPR dataset. This is done by choosing a different subset based on the main source of variation at each bootstrap. We use 4000 bootstrap samples and report 95\% confidence intervals with Bonferroni correction for MoA enrichment on cpg-MoA (Appendix Figures~\ref{fig:boot-moa-fs} and \ref{fig:boot-moa-or}), gene pathway enrichment on cpg-CRISPR (Appendix Figures~\ref{fig:boot-crispr-fs} and \ref{fig:boot-crispr-or}), and replicate retrieval on cpg-tgt2 (Appendix Figures~\ref{fig:boot-tgt2-p1}, \ref{fig:boot-tgt2-map}, and \ref{fig:boot-tgt2-f1}). Under this resampling, the strongest methods (DINOv2, ResNet, and CellProfiler) have overlapping intervals in most settings, whereas the untrained ResNet, and CLOOME for replicate retrieval, perform worse.

\begin{figure}[H]
\centering
\includegraphics[width=\textwidth]{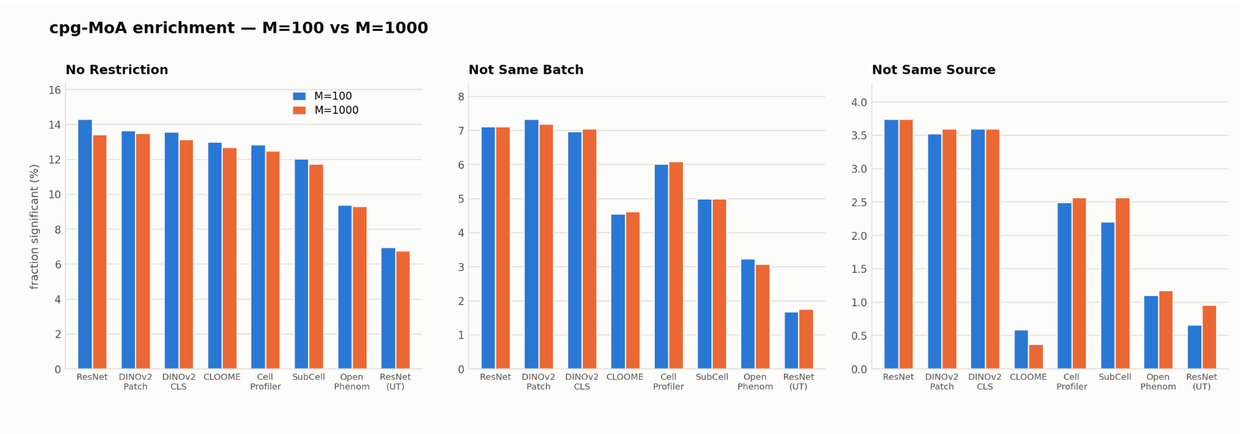}
\caption{Effect of the number of permutations. Fraction significant (\%) for MoA enrichment on cpg-MoA with $M=100$ (as in the main results) and $M=1000$ permutations.}
\label{fig:perm-1000}
\end{figure}

\clearpage
\paragraph{Number of permutations.} The fraction significant metric relies on a permutation test (\hyperref[app:moa-metric]{Appendix A.2}). Increasing the number of permutations from $M=100$ to $M=1000$ changes fraction significant only slightly on cpg-MoA and largely preserves the ranking of models (Appendix Figure~\ref{fig:perm-1000}).

\end{document}